%% file: acl_latex.tex
\documentclass[11pt]{article}

\usepackage[preprint]{acl}

\usepackage{times}
\usepackage{latexsym}

\usepackage[T1]{fontenc}

\usepackage[utf8]{inputenc}

\usepackage{microtype}

\usepackage{inconsolata}

\usepackage{graphicx}
\usepackage{algorithm}
\usepackage{algorithmic}
\usepackage{amsthm}
\usepackage{amsmath}
\usepackage{booktabs}
\usepackage{multirow}
\usepackage{multicol}
\usepackage[table]{xcolor} 
\usepackage[most]{tcolorbox}

\title{Decide Then Retrieve: A Training--Free Framework with Uncertainty--Guided Triggering and Dual-Path Retrieval}

\author{
 \textbf{Wang Chen\textsuperscript{1,2}},
 \textbf{Guanqiang Qi\textsuperscript{1}},
 \textbf{Weikang Li\textsuperscript{3}},
 \\
 \textbf{Yang Li\textsuperscript{1}},
 \textbf{Deguo Xia\textsuperscript{1}},
 \textbf{Jizhou Huang\textsuperscript{1}},
\\
\\
 \textsuperscript{1}Baidu Inc,
 \textsuperscript{2}The University of Hong Kong,
 \textsuperscript{3}Peking University
\\
 \small{
   \textbf{Correspondence:} \href{mailto:email@domain}{liyang164@baidu.com}
 }
}

\begin{document}
\maketitle
\begin{abstract}
Retrieval--augmented generation (RAG) enhances large language models (LLMs) by incorporating external knowledge, but existing approaches indiscriminately trigger retrieval and rely on single-path evidence construction, often introducing noise and limiting performance gains. In this work, we propose \emph{Decide Then Retrieve} (DTR), a training-free framework that adaptively determines \emph{when} retrieval is necessary and \emph{how} external information should be selected. DTR leverages generation uncertainty to guide retrieval triggering and introduces a dual-path retrieval mechanism with adaptive information selection to better handle sparse and ambiguous queries. Extensive experiments across five open-domain QA benchmarks, multiple model scales, and different retrievers demonstrate that DTR consistently improves EM and F1 over standard RAG and strong retrieval-enhanced baselines, while reducing unnecessary retrievals. The code and data used in this paper are available at \url{https://github.com/ChenWangHKU/DTR}.
\end{abstract}

\section{Introduction}

Large language models (LLMs) have dramatically advanced natural language processing (NLP), achieving comparable or even better performance than human beings \cite{touvron2023llama, achiam2023gpt, guo2025deepseek, yang2025qwen3}. However, previous studies \cite{ji2023survey} found that LLMs can only answer questions or accomplish tasks by leveraging their parametric knowledge, which cannot update the latest information nor private datasets. To address this issue, a retrieval--augmented generation (RAG) framework was designed to enhance the generation performance of LLMs using the retrieved information \cite{lewis2020retrieval, guu2020retrieval, karpukhin2020dense, chen2025cmrag}. The effectiveness of RAG has been validated across various NLP tasks, achieving impressive improvement compared with pure LLM systems \cite{ram2023context, gao2023retrieval}.

\begin{figure*}[!ht]
\centering
\includegraphics[width=2.0\columnwidth]{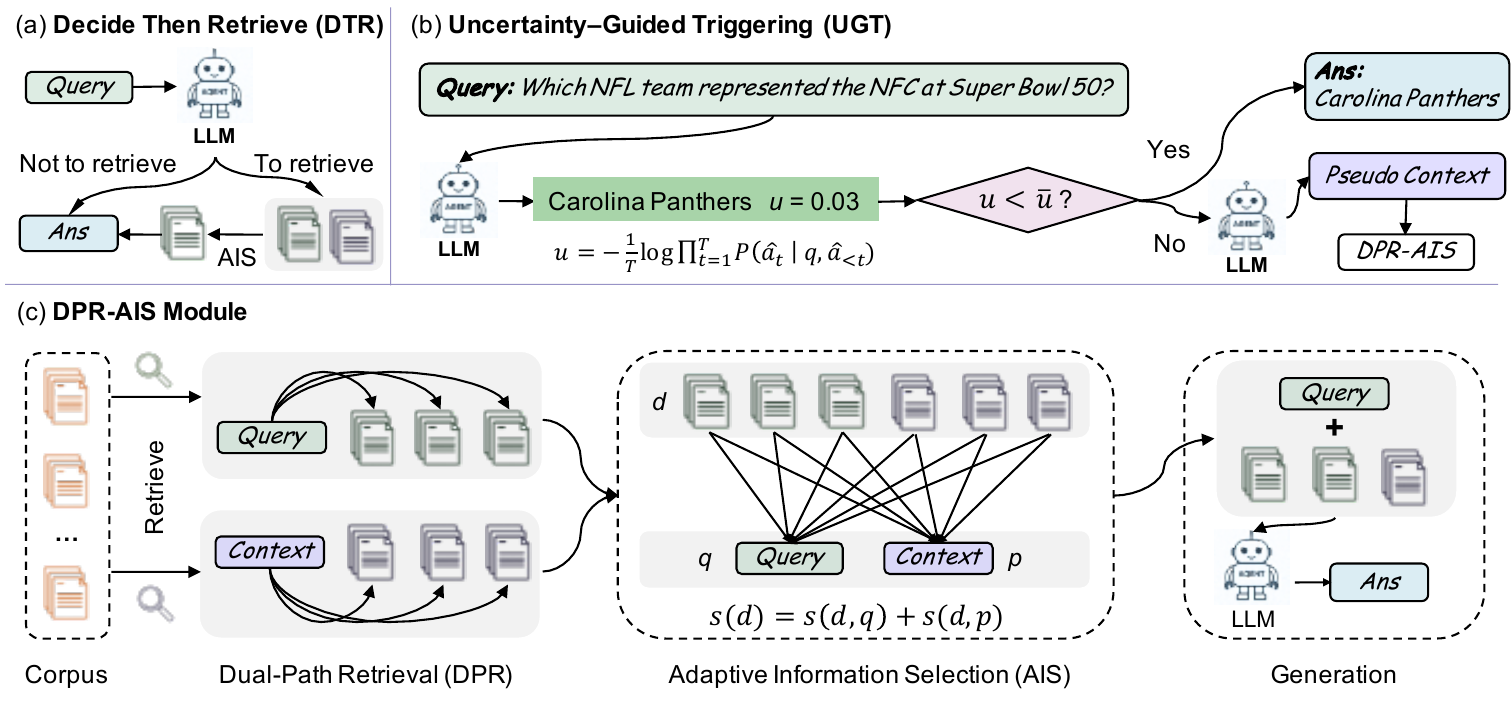}
\caption{Overview of the proposed decide then retrieve (DTR) framework. (a) DTR can adaptively determine whether to retrieve and how to select external information. (b) DTR guides whether to activate retrievals based on the uncertainty score. (c) DTR adaptively selects effective information based on the dual-path retrieval for the final generation.}
\label{fig: framework}
\end{figure*}

Despite its widespread adoption, conventional RAG systems suffer from two critical shortcomings. First, they trigger retrieval for every query, including straightforward questions that could be resolved solely using the LLM’s internal parametric knowledge, resulting in unnecessary noise, which may degrade the final generation accuracy. Second, they exhibit vulnerability to sparse queries---concise user inputs with limited contextual signals---which often yield irrelevant or low-quality retrievals due to inadequate semantic cues, ultimately degrading answer accuracy and reliability \cite{wang2023query2doc, gao2023precise, zhao2024retrieval, singh2025agentic, chen2025pairs}. Please refer to the Appendix~\ref{seca: retrieval_analysis} for detailed analysis.

Existing work \cite{wang2023self, jeong2024adaptive, zhang2025kbm} trains a model to determine whether to retrieve, which may not be generalized to other scenarios, e.g., with different generators. In addition, existing approaches use query augmentation (e.g., appending LLM--generated pseudo-documents or external data to enrich sparse queries) \cite{wang2023query2doc, jagerman2023query, buss2023generating, jeong2024database} to enhance retrieval accuracy. This may induce unnecessary noise due to irrelevant retrievals, degrading the performance of RAG systems in some scenarios.

To address these issues, we introduce \emph{Decide Then Retrieve} (DTR), a unified, training-free framework that rethinks retrieval as a conditional and adaptive process. DTR is built on two key ideas. First, \emph{uncertainty-guided triggering} (UGT) leverages the LLM’s own generation uncertainty to decide whether retrieval is beneficial, allowing confident queries to bypass retrieval and preventing noise from irrelevant evidence. Second, \emph{dual-path retrieval with adaptive information selection} (DPR-AIS) treats the original query and an LLM-generated pseudo-context as complementary signals, retrieving evidence from both perspectives and selecting documents that are jointly relevant.

To summarize, this paper makes the following key contributions:
\begin{itemize}
    \item We propose \emph{Decide Then Retrieve} (DTR), a training-free and model-agnostic RAG framework that can be easily plugged into existing systems, enabling seamless deployment while consistently improving RAG performance.
    \item We introduce uncertainty-guided triggering to selectively activate retrieval based on generation uncertainty and a dual-path retrieval mechanism with adaptive information selection to improve evidence quality for sparse queries.
    \item  We provide a principled accuracy and uncertainty analysis that explains when retrieval is beneficial and how retrieval noise affects generation.
    \item Extensive experiments across five QA datasets, various model sizes, multiple retrieval depths, and different retrievers demonstrate consistent and robust improvements over strong RAG baselines.
\end{itemize}

\section{Related Work}

In this section, we review the related works in recent years, primally focusing on adaptive retrieval and query augmentation.

\subsection{Adaptive retrieval}

\citet{jeong2024adaptive} trained a classifier (i.e., a small LM) to predict the complexity level of incoming queries, achieving a dynamic RAG system. Specifically, the classifier splits queries into three complexity levels: A, B, and C levels, which correspond to \textit{Non Retrieval}, \textit{Single-step Retrieval}, and \textit{Multi-step Retrieval}, respectively, offering an adaptive retrieval approach. However, the classification results may differ from the capabilities of LLMs, which may not extend to other RAG systems. Similarly, a few studies also trained a classifier to predict whether LLMs know the given queries \cite{wang2023self, zhang2025kbm} or each unit’s importance of queries \cite{jia2025find}. Also, a few studies dynamically activated web searching engines when LLMs generated low-confidence tokens \cite{jiang2023active, su2403dragin}. However, these methods may not be generalized to other scenarios, where the classifier or generator is different.

\subsection{Query augmentation}

A query from users may be concise and contain relatively sparse information, which can result in suboptimal retrievals and poor answers. To address this issue, many studies have proposed to augment the query by appending additional terms extracted from retrieved documents \cite{lavrenko2017relevance, lv2009comparative} or generated by neural models \cite{zheng2020bert, mao2021generation}. Recently, many studies leveraged advanced LLMs to augment queries and achieved notable improvement \cite{buss2023generating, wang2023query2doc, jagerman2023query, gao2023precise, lin2023train}. For example. \citet{wang2023query2doc} first prompted LLMs to generate a pseudo-document of the query, which was then concatenated with the original query to enhance the retrieval. In addition, a few studies \cite{jeong2022augmenting, jeong2024database} used an external database, such as relevant tabular data, to augment original queries, which can enhance query representations but require additional data. Furthermore, a few studies \cite{chan2024rq, zhang2025imprag} trained models to refine or encode queries to enhance the performance of information retrieval and question answering. However, these

\section{Preliminaries}

In this section, we first define the RAG system, and then, we analyze the primary facts that affect the generation accuracy of a RAG system.

\subsection{Problem setup}

In a retrieval--augmented generation (RAG) system, a collection of documents is first split into small chunks $D$, which are then embedded into vectors using an embedding model $f$. Given a query $q$, the retriever $\mathcal{R}$ first searches top $k$ relevant chunks $D_k = \{d_1, d_2, ..., d_k\} \subset D$ from the collection according to the similarity between the query and chunks, i.e., $D_k = \mathcal{R}(D, q)$. The similarity $s$ is calculated using the inner product (IP) or L2 norm between the query embedding $\mathbf{q} = f(q)$ and chunk embeddings $\mathbf{d} = f(d), \forall d \in D$. Finally, the retrievals and query are incorporated into a prompt $\mathcal{P}(q, D_k)$, which is used as the input of the generator model $g$ to generate the final answer, i.e.,  $\hat{a} = \mathcal{G}(\mathcal{P}(q, D_k))$. We list all notations and abbreviations in Appendix \ref{seca: notation}.

\subsection{Accuracy analysis of RAG}

\begin{figure}[!ht]
\centering
\includegraphics[width=1.0\columnwidth]{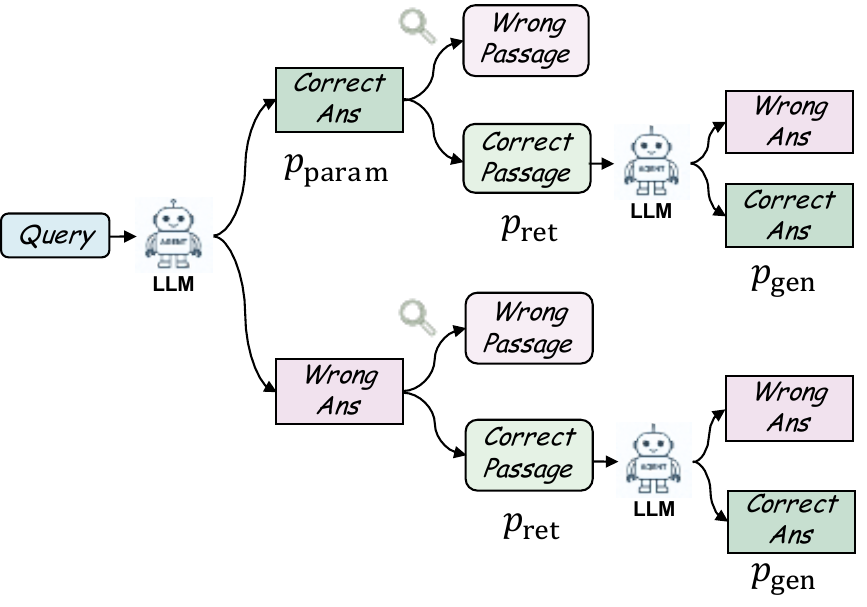}
\caption{Overview of RAG accuracy analysis.}
\label{fig: RAG_acc_analysis}
\end{figure}

Given a query $q$, we analyze the probability that a retrieval-augmented generation (RAG) system produces an accurate answer by decomposing the process into parametric answering, retrieval, and generation stages, as illustrated in Figure~\ref{fig: RAG_acc_analysis}. We denote by $P_{\text{param}}$ the probability that the large language model (LLM) can directly generate a correct answer based solely on its parametric knowledge, without relying on external evidence. When $P_{\text{param}}$ is low, the system resorts to external retrieval. Let $P_{\text{ret}}$ denote the probability that the retriever returns correct and relevant passages for the query, capturing the effectiveness of the retrieval module. Given correct retrievals, the probability that the LLM successfully generates an accurate answer from the retrieved passages is denoted as $P_{\text{gen}}$, which reflects the model’s robustness to noise and its long-context reasoning capability.

We assume that the LLM cannot generate an accurate answer when conditioned on incorrect retrievals. Under this assumption, the overall probability of obtaining a correct answer via retrieval is $P_{\text{ret}} \cdot P_{\text{gen}}$. Therefore, external retrieval should be triggered only when retrieval-based answering is expected to outperform parametric answering, i.e.,
\begin{equation}
P_{\text{ret}} \cdot P_{\text{gen}} > P_{\text{param}} .
\end{equation}
This formulation yields several insights. First, when $P_{\text{param}}$ is sufficiently high, incorporating retrieval may degrade performance due to imperfect retrieval accuracy. Second, for a fixed LLM, both $P_{\text{param}}$ and $P_{\text{gen}}$ are primarily determined by the model’s intrinsic capability, whereas $P_{\text{ret}}$ depends on the retrieval system. Consequently, we identify \textbf{two key research questions:} (1) \textit{how to determine, for a given query, whether retrieval should be invoked;} and (2) \textit{how to improve retrieval accuracy to maximize the overall effectiveness of RAG systems.}

\section{Decide Then Retrieve}

In this section, we introduce the proposed decide then retrieve (\textbf{DTR}) method, including (1) uncertainty--guided triggering (\textbf{UGT}) and (2) dual-path retrieval with adaptive information selection (\textbf{DPR-AIS}).

\subsection{Uncertainty--Guided Triggering}
\label{subsec:uncertainty_trigger}

Given a query $q$, the LLM generates an answer $\hat{a}=\{\hat{a}_1,\dots,\hat{a}_T\}$ in an autoregressive manner, where the generation probability is
\begin{equation}
    P(\hat{a} \mid q) = \prod_{t=1}^{T} P(\hat{a}_t \mid q, \hat{a}_{<t}) .
\end{equation}
Following recent work~\cite{kang2025scalable, fu2025deep}, we define the uncertainty of the generated answer as the normalized negative log-likelihood:
\begin{equation}
    u = - \frac{1}{T} \log P(\hat{a} \mid q).
\end{equation}
This uncertainty score measures the model’s confidence in its own prediction and can be computed directly from next-token probabilities.

\begin{figure}[!ht]
\centering
\includegraphics[width=1.\columnwidth]{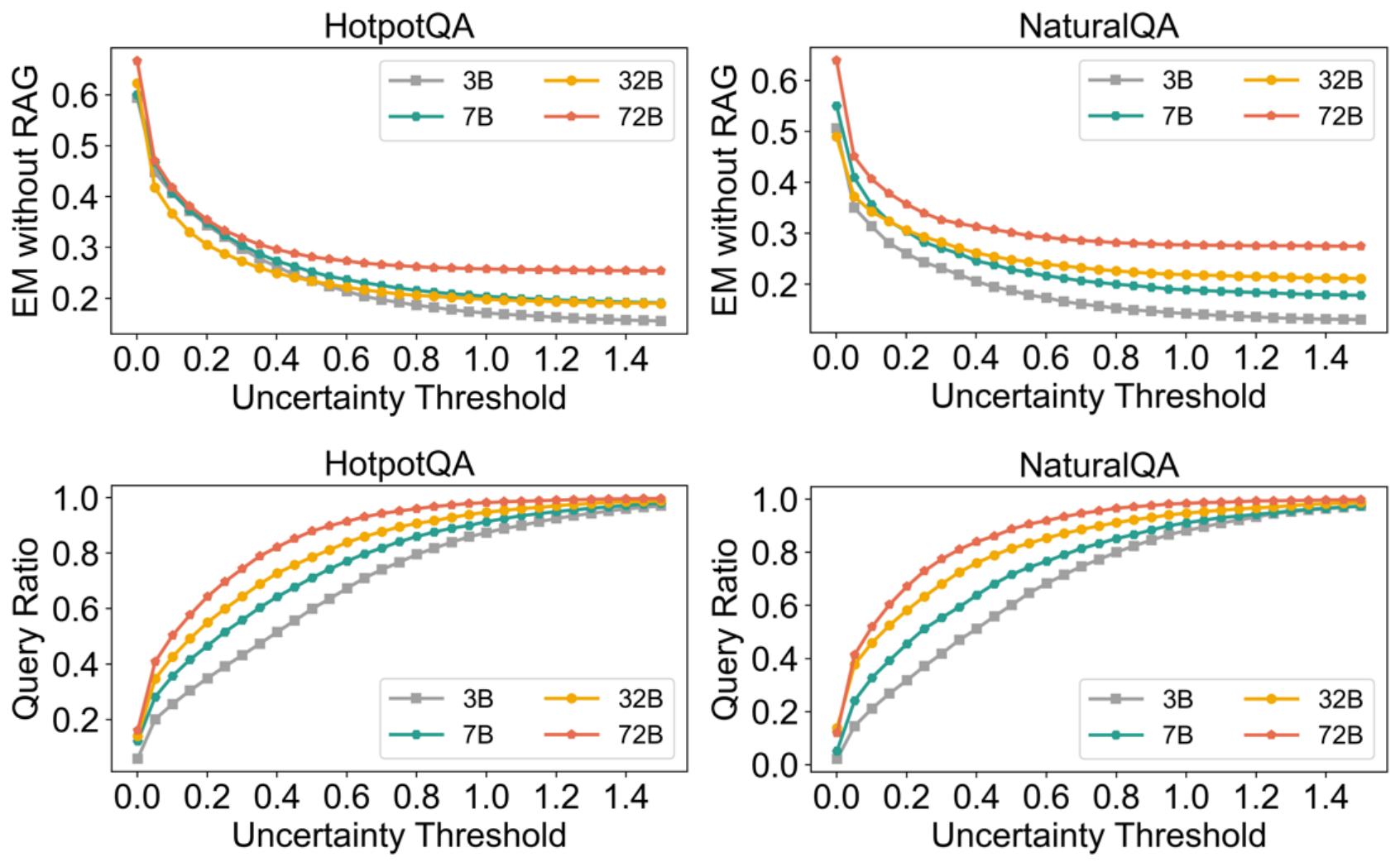}
\caption{Generation Uncertainty vs. Parametric Accuracy and Query Coverage. \textit{Qwen2.5} series models are used as the generators.}
\label{fig: EM_uncertainty}
\end{figure}

Figure~\ref{fig: EM_uncertainty} analyzes how uncertainty relates to answer accuracy and retrieval triggering across HotpotQA~\cite{yang2018hotpotqa} and NaturalQA~\cite{kwiatkowski2019natural} (please refer to Appendix~\ref{seca: additional_uncertainty_measure} for more results). As the uncertainty threshold increases, the exact match (EM) score of answers generated \emph{without} retrieval monotonically decreases, indicating that higher uncertainty is strongly associated with lower parametric accuracy. This indicates that the proposed uncertainty score can be used to guide the decision of retrievals. At the same time, the query ratio---defined as the proportion of queries whose uncertainty is no greater than the threshold---monotonically increases, implying that a higher threshold allows more queries to bypass retrieval.

\subsection{Dual-path retrieval}

\begin{figure}[!ht]
\centering
\includegraphics[width=1.0\columnwidth]{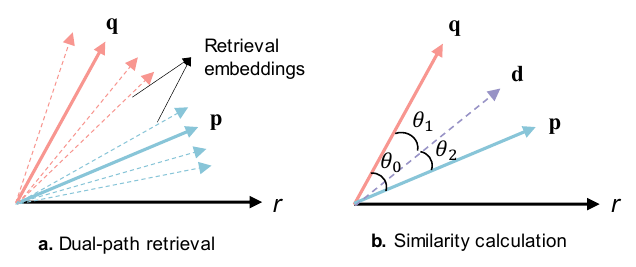}
\caption{Illustration of (a) dual-path retrieval mechanism and (b) similarity calculation of retrievals.}
\label{fig: retrieval}
\end{figure}

Unlike prior approaches that first generate a pseudo-context (e.g., an answer or rationale) and then merge it with the original query into a single retrieval signal—an operation that can amplify noise due to LLM hallucinations—we adopt a \emph{dual-path retrieval} (DPR) mechanism that treats the query and the self-generated context as complementary but independent sources of information. Specifically, DPR executes two parallel retrieval operations---one conditioned on the query and the other on the pseudo-context. Specifically, as shown in Figure \ref{fig: retrieval} (a), the system retrieves top $n$ relevant documents using the embeddings of the query $\mathbf{q} = f(q)$ and the generated context $\mathbf{p} = f(p)$, respectively, and the retrieved documents can be denoted as $D_{2n} = \{d_1, d_2, ..., d_{2n}\} \subset D$. This dual-path strategy enhances the relevance and diversity of the retrieved documents, which is especially beneficial for complex or ambiguous queries where either source alone may fall short.

\paragraph{Adaptive information selection.}

Once the dual-path retrieval yields a set of $2n$ candidate documents based on the original query and the LLM-generated pseudo-context, we introduce an adaptive information selection (AIS) mechanism to refine this candidate set. The goal is to prioritize documents that are simultaneously relevant to $q$ and $p$, ensuring that the final context passed to the LLM is not only topically aligned but also semantically coherent from different perspectives. Formally, we compute the score for each document $s(d)$ as follows:
\begin{equation}
    s(d) = s_1(d, q) + s_2(d, p), \quad \forall d \in D_{2n},
\end{equation}
where $s_1(d,q)$ and $s_2(d,p)$ denote the relevance between the document $d$ and the query $q$ and LLM--generated context $p$, respectively.

As shown in Figure \ref{fig: retrieval} (b), a straightforward method is to calculate the relevance ($s_1$ and $s_2$) as the IP between their corresponding embeddings, as follows:
\begin{equation}
    s_1 (d, q) = \cos (\theta_1) = \langle \mathbf{d}, \mathbf{q} \rangle,
\end{equation}
\begin{equation}
    s_2(d, p)= \cos (\theta_2) = \langle \mathbf{d}, \mathbf{p} \rangle,
\end{equation}
where $\mathbf{q} = f(q)$, $\mathbf{p} = f(p)$, and $\mathbf{d} = f(d), \forall d \in D_{2n}$. Remind that $\mathbf{q}$, $\mathbf{p}$, and $\mathbf{d}$ are normalized vectors. However, such aggregation may not fully capture the joint relevance, especially if either $q$ or $p$ introduces semantic noise. Therefore, AIS leverages a geometric intuition: if both $\mathbf{q}$ and $\mathbf{p}$ align well with a given document $\mathbf{d}$, then the combined angle $\theta = \theta_1 + \theta_2$ should be minimized. This motivates the maximization of $s(d) = \cos(\theta) = \cos(\theta_1 + \theta_2)$, which expands to:
\begin{align}\label{eq: score}
    & s(d) = \cos (\theta_1 + \theta_2) \nonumber \\
    & = s_1 \cdot s_2 - \sqrt{1-(s_1)^2} \cdot \sqrt{1-(s_2)^2}.
\end{align}
This formulation implicitly rewards documents that are jointly aligned with both $\mathbf{q}$ and $\mathbf{p}$ while penalizing those that diverge from either direction. The DPR-AIS algorithm, i.e., dual-path retrieval--based adaptive information selection, is formulated as Algorithm \ref{alg: DPR-AIS}.

\begin{algorithm}[!htb]
\caption{DPR-AIS}
\label{alg: DPR-AIS}
\textbf{Input}: Query $q$, LLM--generated pseudo-context $p$,  encoder $f(\cdot)$, retriever $\mathcal{R}$, corpus $D = \{d_1, d_2, \dots, d_N\}$, retrieval size $n$, and selection size $k$\\
\textbf{Output}: Filtered relevant documents $D_k \subset D$

\begin{algorithmic}[1]
\STATE Compute embeddings: $\mathbf{q} \leftarrow f(q),\; \mathbf{p} \leftarrow f(p),\; \mathbf{d} \leftarrow f(d)$
\STATE Retrieve top-$n$ documents using $\mathbf{q}$: $D_q \leftarrow \mathcal{R}(D, \mathbf{q})$
\STATE Retrieve top-$n$ documents using $\mathbf{p}$: $D_p \leftarrow \mathcal{R}(D, \mathbf{p})$
\STATE Merge retrieved sets: $D_{2n} \leftarrow D_q \cup D_p$
\FOR{each $d \in D_{2n}$}
    \STATE Compute $s_1 = \langle \mathbf{q}, \mathbf{d} \rangle$ and $s_2 = \langle \mathbf{p}, \mathbf{d} \rangle$
    \STATE Compute joint relevance score using Eq. \eqref{eq: score}
\ENDFOR
\STATE Select top-$k$ documents from $D_{2n}$ by descending $s(d)$ values
\STATE \textbf{return} selected document set $D_k$
\end{algorithmic}
\end{algorithm}

\section{Experiments}

\subsection{Experiment setup}

\paragraph{Evaluation datasets.}
We evaluate the effectiveness of the proposed method on open domain QA datasets: (1) NaturalQA \cite{kwiatkowski2019natural}, (2) WebQuestions \cite{berant2013semantic}, (3) SQuAD \cite{rajpurkar2016squad}, and (4) TriviaQA \cite{joshi2017triviaqa}. In addition, we further test the method using one complex multi-hop QA dataset—HotpotQA \cite{yang2018hotpotqa}. For the open domain datasets, we use the 21M English Wikipedia dump as the corpus for retrieval, while for HotpotQA, we use its original corpus. The statistics of all datasets are listed in Table \ref{tab: data}.

\begin{table}[!ht]
\centering
\begin{tabular}{l l l}
\toprule
    Datasets & \#Questions & \#Documents \\
\midrule
    NaturalQA    & 3,610  & 21 M \\
    WebQuestions & 2,032  & 21 M \\
    SQuAD        & 10,570 & 21 M \\
    TriviaQA     & 11,313 & 21 M \\
    HotpotQA        & 7,405 & 5 M \\
\bottomrule  
\end{tabular}
\caption{Statistics of evaluation datasets.}
\label{tab: data}
\end{table}

\paragraph{Baselines.}
We compare the proposed method with other training--free baselines. (1) \textbf{No Retrieval} leverages the parametric knowledge of LLM to directly generate a response to the given query without retrieval. (2) \textbf{Standard RAG} incorporates the retrieved top-$k$ documents with the query as the input of LLM through prompting. (3) \textbf{LLM Judge} \cite{wang2023self, jeong2024adaptive, zhang2025kbm} determines whether to trigger retrievals based on the LLM. (4) \textbf{HyDE} \cite{gao2023precise} leverages the LLM--generated passage to enhance the retrieval. (5) \textbf{Q2D} \cite{wang2023query2doc} concatenates multiple queries and the LLM--generated pseudo-document to perform the retrieval. (6) \textbf{CoT} \cite{jagerman2023query} first prompts the LLM to generate the answer as well as the rationale to the given query, and then combines multiple queries and the LLM outputs into the retrieval signal.

\paragraph{Evaluation metrics.}
To comprehensively assess the quality of generated answers, we adopt two widely used metrics in open-domain question answering: Exact Match (EM) and F1 score. The EM metric quantifies the proportion of predictions that exactly match any one of the ground-truth answers, serving as a strict measure of correctness. On the contrary, the F1 score provides a more forgiving evaluation by computing the token-level overlap between the predicted and reference answers. We normalize the predicted and ground truth answers following the implementation of \citet{fang2025kirag}.

\input{tables/main_results}

\paragraph{Implementation details.}
We implement our DTR framework using Qwen2.5 series models~\cite{team2024qwen2} as the backbone LLM for answer and context generation and retrieval triggering judge. To ensure deterministic outputs and eliminate variability due to random sampling, we set the temperature to 0.0 during decoding \cite{kimsure}. For dense retrieval, we use the bge-large-en-v1.5 embedding model~\cite{bge_embedding} and e5~\cite{wang2022text}, employing the inner product (IP) as the similarity metric. In the dual-path retrieval step, we retrieve the top 5 most relevant documents independently for both the query and the generated pseudo-context, resulting in a combined candidate pool of 10 documents ($2n = 10$). All remaining experimental configurations of baselines follow the implementations reported in their respective original papers. We present the prompts used in this study in Appendix~\ref{seca: prompt}.

\input{tables/ablation_results}

\subsection{Main results}

\paragraph{Overall performance.}
Table~\ref{tab: results} summarizes the main results on five QA benchmarks using \textit{Qwen2.5-7B-Instruct} and \textit{Qwen2.5-72B-Instruct} as generators. Across both model scales and all datasets, DTR consistently achieves the best or second-best performance in terms of both EM and F1. For the 7B model, DTR improves the average EM/F1 from 35.81/45.81 (standard RAG) to up to 37.87/48.08, demonstrating clear gains over static retrieval. Similar improvements are observed for the 72B model, where DTR reaches an average EM/F1 of 40.46/52.14, outperforming standard RAG and all competing baselines. We list more evaluation results with top-5 retrievals or e5 as the retriever in Appendix~\ref{seca: additional_results}.

\paragraph{Effectiveness of uncertainty-guided triggering.}
Compared to standard RAG, which triggers retrieval for all queries, DTR selectively activates retrieval based on uncertainty, which yields higher accuracy, confirming that unnecessary retrievals can introduce noise and degrade generation quality. On the contrary, LLM Judge with a small model (7B) rarely triggers retrieval, resulting in performance nearly identical to the no-retrieval baseline, while LLM Judge (72B) triggers retrieval frequently but remains inferior to DTR, highlighting the advantage of uncertainty-based signals over heuristic or judge-based decisions. In addition, varying the uncertainty threshold $u$ allows DTR to trade off retrieval coverage and accuracy. Smaller thresholds lead to higher trigger ratios and slightly better performance on reasoning-intensive datasets (e.g., HotpotQA, NaturalQA), while larger thresholds reduce retrieval frequency with marginal accuracy loss. This flexibility enables DTR to adapt to different deployment constraints.

\paragraph{Comparison with retrieval-enhanced prompting.}
Methods such as HyDE, Q2D, and CoT improve over standard RAG in some cases but are consistently outperformed by DTR. This indicates that the proposed dual-path retrieval mechanism with adaptive information selection is more robust to noise and can retrieve more relevant evidence, upgrading the overall performance of RAG systems.


\subsection{Ablation results}

Table~\ref{tab: ablation_results} presents ablation studies that analyze the contribution of each component in DTR under both \textit{Qwen2.5-7B-Instruct} and \textit{Qwen2.5-72B-Instruct}.

\paragraph{Effect of uncertainty-guided triggering (UGT).}
Removing UGT results in comparable or slightly lower scores, indicating that UGT effectively avoids activating unnecessary retrievals without sacrificing overall accuracy. Notably, the performance gap is more evident for the 72B model, suggesting that stronger parametric capability benefits more from uncertainty-guided triggering, as accurate answers can be generated directly without relying on external evidence.

\paragraph{Effect of dual-path retrieval (DPR).}
Disabling DPR (\textit{w/o DPR}) consistently degrades performance across datasets and model scales. For the 7B model, the average EM/F1 drops from 37.87/48.08 to 35.98/45.86, and similar declines are observed for the 72B model. This confirms that the dual-path retrieval mechanism is a critical component, enabling complementary evidence acquisition that improves answer generation.

\input{tables/retrieval_acc}

\paragraph{Effect of adaptive information selection (AIS).}
We further ablate AIS by fixing the composition of retrieved information. When AIS is removed, and the system uses static combinations (1$q$+2$p$ or 2$q$+1$p$), performance drops substantially, especially for the 7B model. This demonstrates that adaptively selecting and balancing different retrieval paths is essential for effectively leveraging retrieved evidence.


\subsection{Analysis}

\paragraph{Retrieval accuracy.}

Table~\ref{tab: retrieval_acc} compares retrieval accuracy on HotpotQA using two retrievers, \textit{bge} and \textit{e5}. Among all methods, DPR achieves the highest retrieval accuracy under both retrievers. Specifically, DPR improves retrieval accuracy from 61.9\% to 62.7\% with \textit{bge}, and from 59.3\% to 62.6\% with \textit{e5}. On the contrary, query expansion methods such as HyDE, Q2D, and CoT consistently underperform standard RAG, indicating that naive expansion may introduce noise and dilute retrieval relevance. These results demonstrate that the proposed dual-path retrieval strategy is more effective at identifying relevant evidence than conventional single-path retrieval or query expansion methods, providing a stronger foundation for downstream generation. It should be noted that we only evaluate retrieval accuracy on HotpotQA, as it is paired with a well-defined corresponding corpus.

\paragraph{Uncertainty scaling.}

We evaluate the EM improvement obtained by activating retrieval relative to no retrieval across different uncertainty thresholds. As shown in Figure~5, retrieval degrades accuracy when uncertainty is low, indicating that unnecessary retrieval introduces noise. As uncertainty increases, the improvement ratio consistently rises, showing that retrieval becomes increasingly helpful when parametric knowledge is insufficient.

\begin{figure}[!ht]
\centering
\includegraphics[width=1.0\columnwidth]{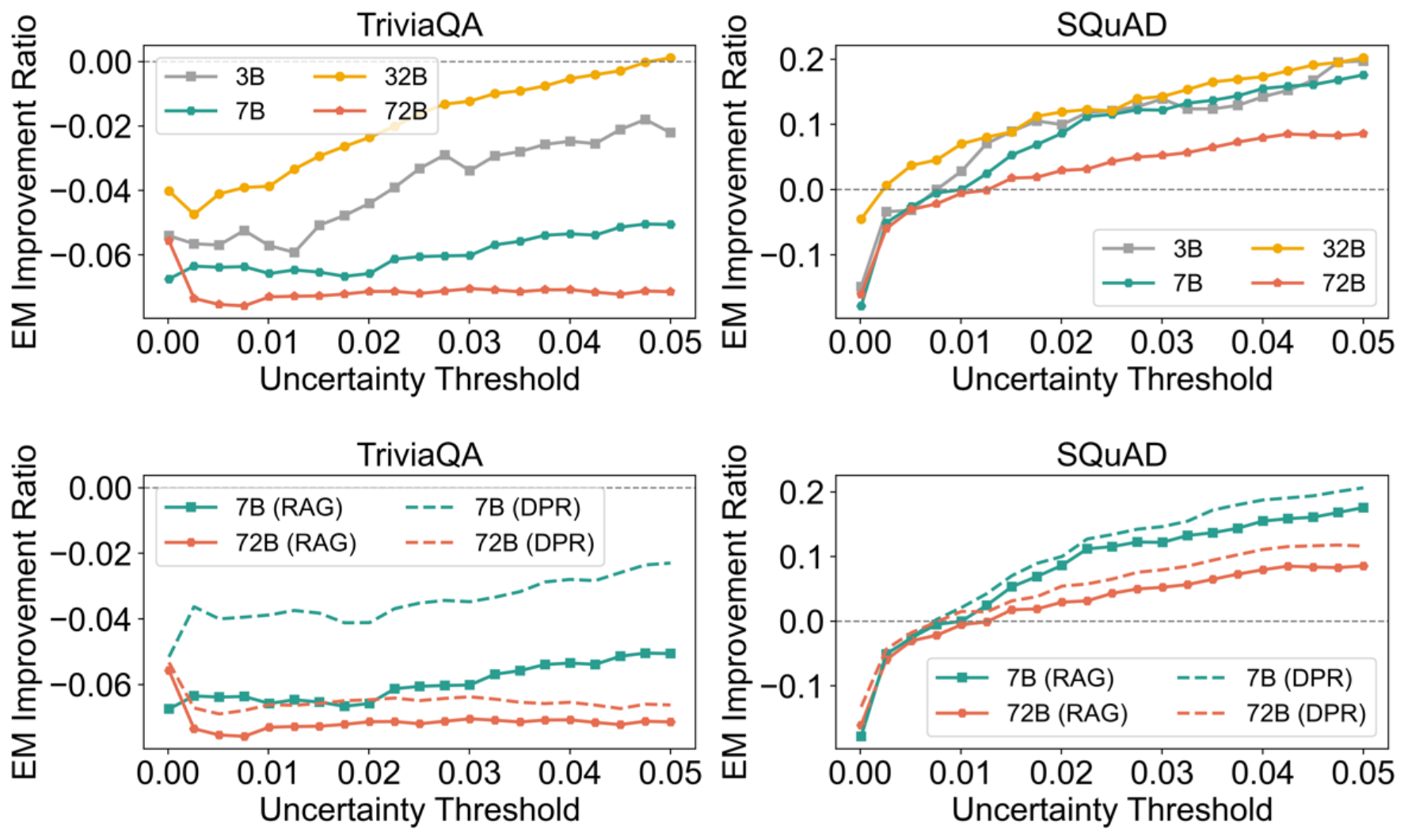}
\caption{Uncertainty scaling results across various model sizes and different retrieval mechanisms.}
\label{fig: uncertainty_scaling}
\end{figure}

\begin{figure}[!ht]
\centering
\includegraphics[width=1.0\columnwidth]{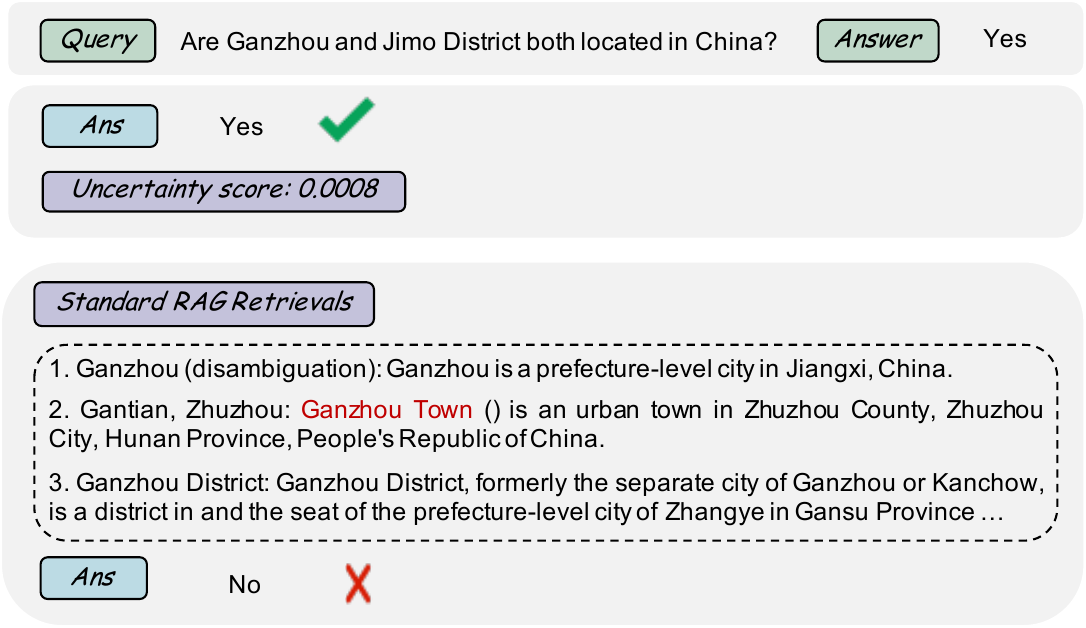}
\caption{Comparison between uncertainty--guided triggering and standard RAG.}
\label{fig: case_study_1}
\end{figure}

Model scale significantly affects this trend. The 72B model exhibits slower gains due to its strong parametric capability, while the 32B model achieves the largest improvements, benefiting from a favorable balance between parametric knowledge and robustness to retrieval noise. In addition, more accurate retrieval mechanisms yield larger improvements under the same uncertainty thresholds, as increased retrieval accuracy (higher $P_{\text{ret}}$) amplifies the effectiveness of retrieval. We report all scaling results in Appendix~\ref{seca: uncertainty_scaling}.



\paragraph{Case study.}
Fig. \ref{fig: case_study_1} demonstrates a representative example. The query is a simple factual comparison that lies well within the LLM's parametric knowledge, and thus, the model accurately generates answers correctly with a lower uncertainty score. However, irrelevant documents retrieved using the query can lead to an inaccurate answer. We present more cases in the appendix~\ref{seca: additional_case_study}.

\section{Conclusion}
This paper presents DTR, a training-free framework that enhances retrieval--augmented generation through uncertainty-guided triggering and dual-path retrieval with adaptive information selection. By explicitly deciding when retrieval is necessary and how to leverage complementary retrieval signals, DTR improves both accuracy and robustness over conventional RAG pipelines. Extensive experiments demonstrate that DTR consistently outperforms strong baselines across model scales, datasets, and retrievers, establishing it as a practical and scalable solution for adaptive RAG systems.

\newpage
\section*{Limitations}
While DTR is effective and lightweight, it relies on the quality of uncertainty estimates derived from token probabilities, which may vary across decoding strategies or model families. In addition, generating pseudo-contexts introduces extra inference steps, increasing latency compared to single-pass retrieval. Finally, although DTR generalizes well across retrievers, its performance still depends on the underlying retrieval corpus and embedding quality.

\section*{Ethical Considerations}
DTR does not introduce new data sources or training procedures beyond standard RAG setups, and thus inherits existing ethical considerations related to LLMs and information retrieval, such as potential biases in corpora and misinformation in retrieved documents. By reducing unnecessary retrievals, DTR may mitigate exposure to irrelevant or misleading content, but it does not eliminate the need for careful dataset curation and responsible deployment.


\bibliography{custom}

\newpage
\appendix

\section{Additional related work}

How to retrieve and select relevant information is a key issue in RAG systems. With the remarkable success of advancing LLMs using reinforcement learning (RL), many studies have adopted RL to train LLMs to enhance information retrieval and reranking \cite{asai2024selfrag, jin2025search, li2025search, song2025r1, yu2024rankrag}. In addition, several studies \cite{jiang2025gainrag, yan2025rpo} have aligned the retriever's preferences with those of LLMs to retrieve more relevant information and enhance generation performance. These methods could achieve significant improvement, but may suffer from high computational costs. In addition, with the accessibility to advanced reranking models, such as bge-reranker \cite{bge_embedding} and Qwen3-reranker \cite{zhang2025qwen3}, a straightforward method is first to rerank the retrievals using a reranker and then select the top-$k$ relevant retrievals \cite{glass2022re2g, chang-etal-2025-main}. This method could further select retrievals with a higher semantic similarity with the query and thus enhance the generation performance. However, it may not perform well or even degrade answer accuracy in QA tasks due to the sparse query or large corpus \cite{kimsure, jiang2025gainrag}.

\section{Notation list}
\label{seca: notation}

The main notations and abbreviations used in this paper are listed in Table \ref{tab: notation}.

\begin{table}[!ht]
\centering
\resizebox{0.99\columnwidth}{!}{
\begin{tabular}{l l}
\toprule
    Notation &  Explanation\\
\midrule
    $P_{\text{param}}$ & Probability of accurate parametric generation \\
    $P_{\text{ret}}$ & Probability of accurate retrieval \\
    $P_{\text{gen}}$ & Probability of accurate final generation \\
    $D$           & Chunks of Documents \\
    $D_k$         & Retrieved $k$ chunks \\
    $f$           & Embedding model \\
    $q$           & Query \\
    $\mathbf{q}$  & Query embedding \\
    $d$           & A chunk \\
    $\mathbf{d}$  & Chunk embedding \\
    $\mathcal{R}$ & Retriever \\
    $\mathcal{P}$ & Prompt \\
    $\mathcal{G}$ & LLM generator \\
    $\hat{a}$     & Generated answer \\
    $p$           & LLM--generated pseudo-context \\
    $\mathbf{p}$  & Pseudo-context embedding \\
    $\theta_0$    & The angle between $\mathbf{q}$ and $\mathbf{p}$ \\
    $\theta_1$    & The angle between $\mathbf{q}$ and $\mathbf{d}$ \\
    $\theta_2$    & The angle between $\mathbf{p}$ and $\mathbf{d}$ \\
    $s(d)$        & Score of $d$ \\
\midrule
    Abbreviation &  Explanation \\
\midrule
    DTR   & Decide then retrieve \\
    UGT   & Uncertainty--guided triggering \\
    DPR   & Dual-path retrieval \\
    AIS   & Adaptive information selction \\
    IP    & Inner product \\
    EM    & Exact match \\

\bottomrule  
\end{tabular}}
\caption{Explanation of main notations and abbreviations used in this study.}
\label{tab: notation}
\end{table}

\section{Retrieval analysis}
\label{seca: retrieval_analysis}

In this section, we further analyze the properties of retrievals using the query or LLM--generated pseudo-context. Specifically, we first compare the documents retrieved using the query with the ground-truth (GT) documents. We found that the retrieved documents may be significantly different from the GT documents. In addition, we analyze the spatial distribution of query--based, pseudo-context--based, and GT documents. We discovered that the documents retrieved using the pseudo-context may compensate for this shortcoming.

\subsection{Similarity between query and GT documents}

\begin{figure}[!ht]
    \centering
    \includegraphics[width=1.0\linewidth]{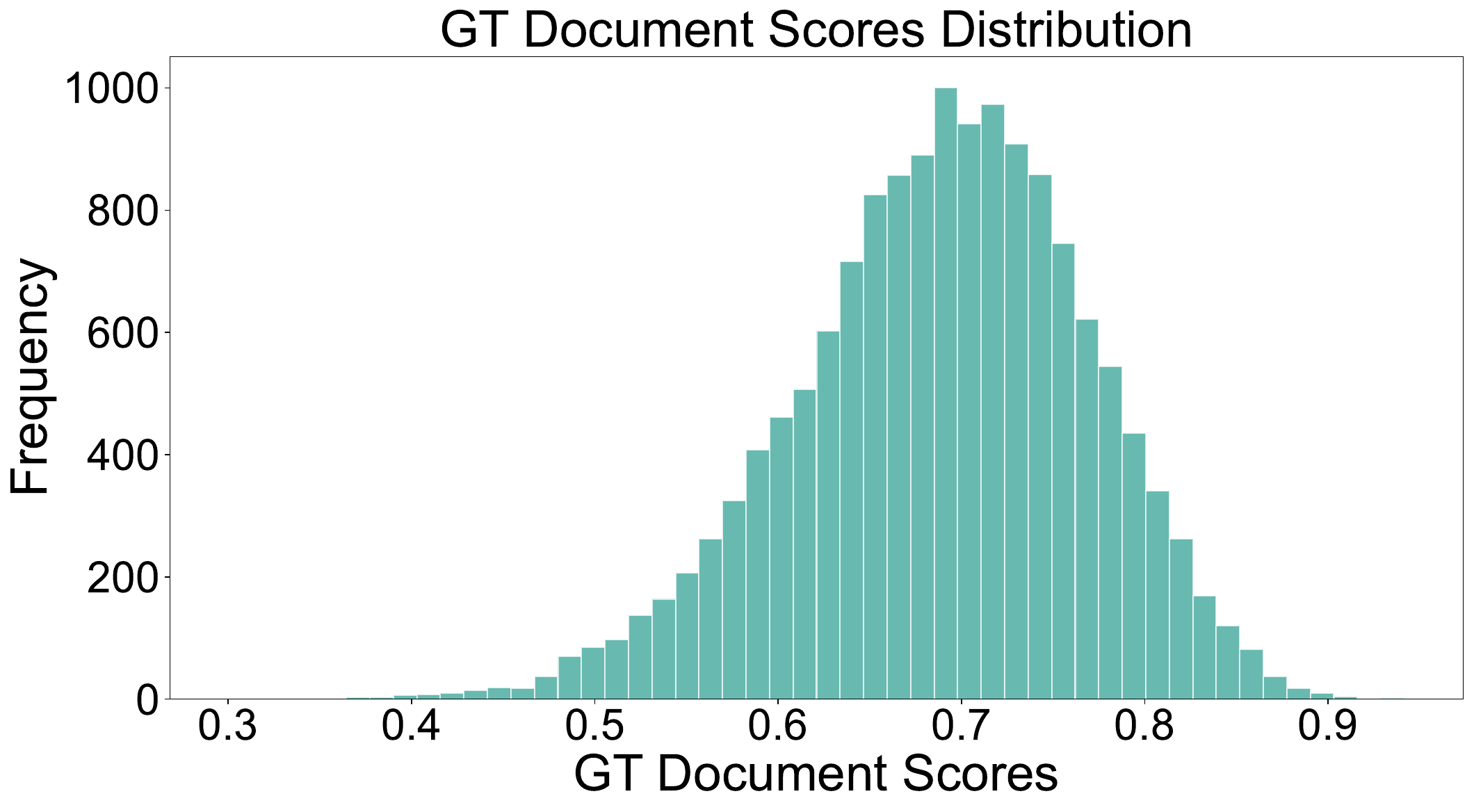}
    \caption{Similarity scores between queries and ground truth documents.}
    \label{fig: GT_doc_score}
\end{figure}

To figure out how the query--based documents differ from the GT documents, we first calculated the similarity scores between the ground-truth(GT) documents and the queries in the HotpotQA dataset. As shown in Fig. \ref{fig: GT_doc_score}, the similarity scores of the most GT documents are below 0.8, indicating that the query may differ from the GT documents, which can lead to irrelevant retrievals.

\begin{figure}[!ht]
    \centering
    \includegraphics[width=1.0\linewidth]{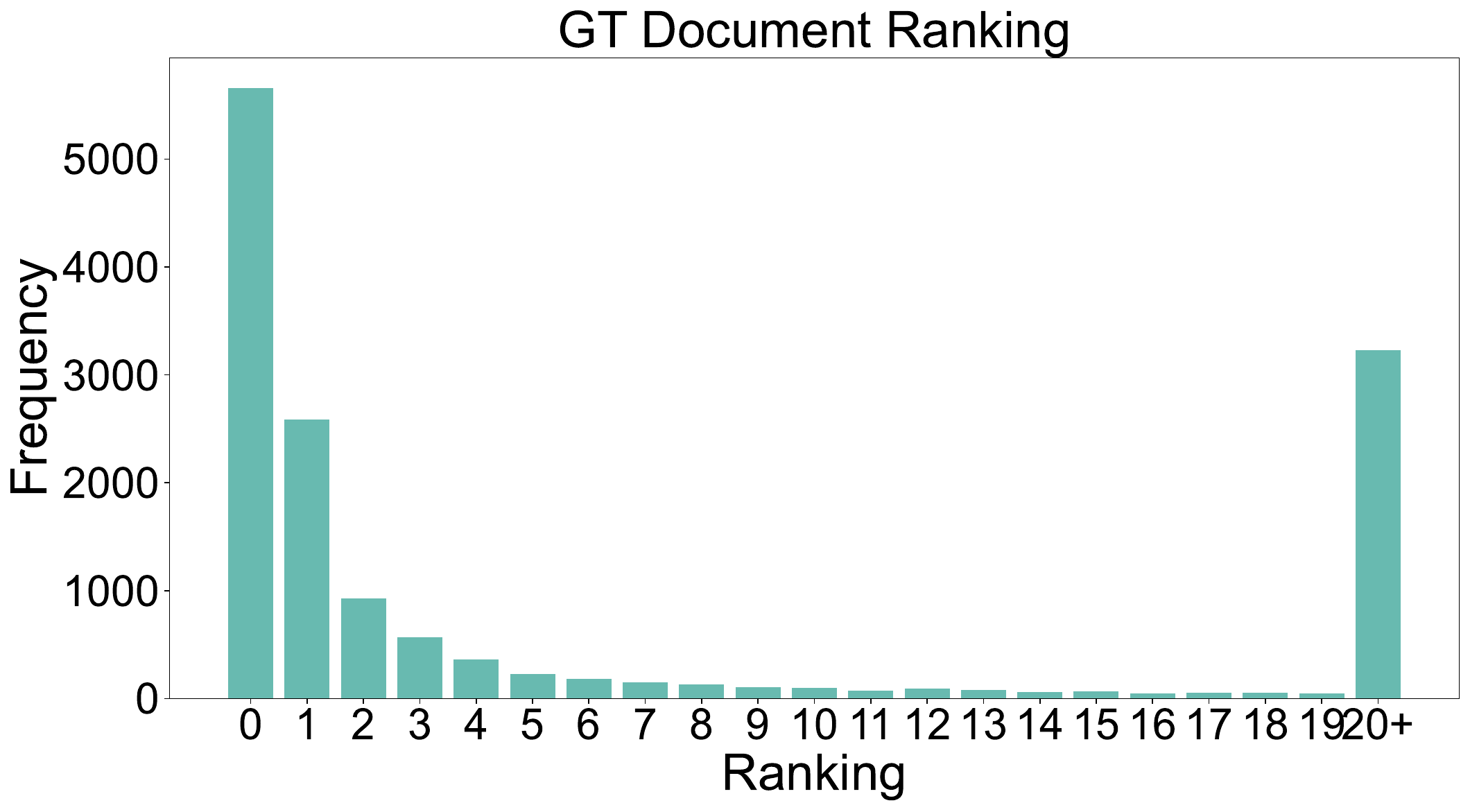}
    \caption{Ranking of ground truth documents in query--based retrievals.}
    \label{fig: GT_doc_orders}
\end{figure}

Furthermore, we ranked the GT documents in the query--based retrievals. As shown in Fig. \ref{fig: GT_doc_orders}, a significant proportion of the GT documents rank 20+ in the documents retrieved using the query. This further demonstrates that using the query to retrieve documents only can lead to irrelevant results, which in turn result in inaccurate final answers.

\subsection{Spatial distribution of query, context, and GT documents}

\begin{figure}[!ht]
    \centering
    \includegraphics[width=0.7\linewidth]{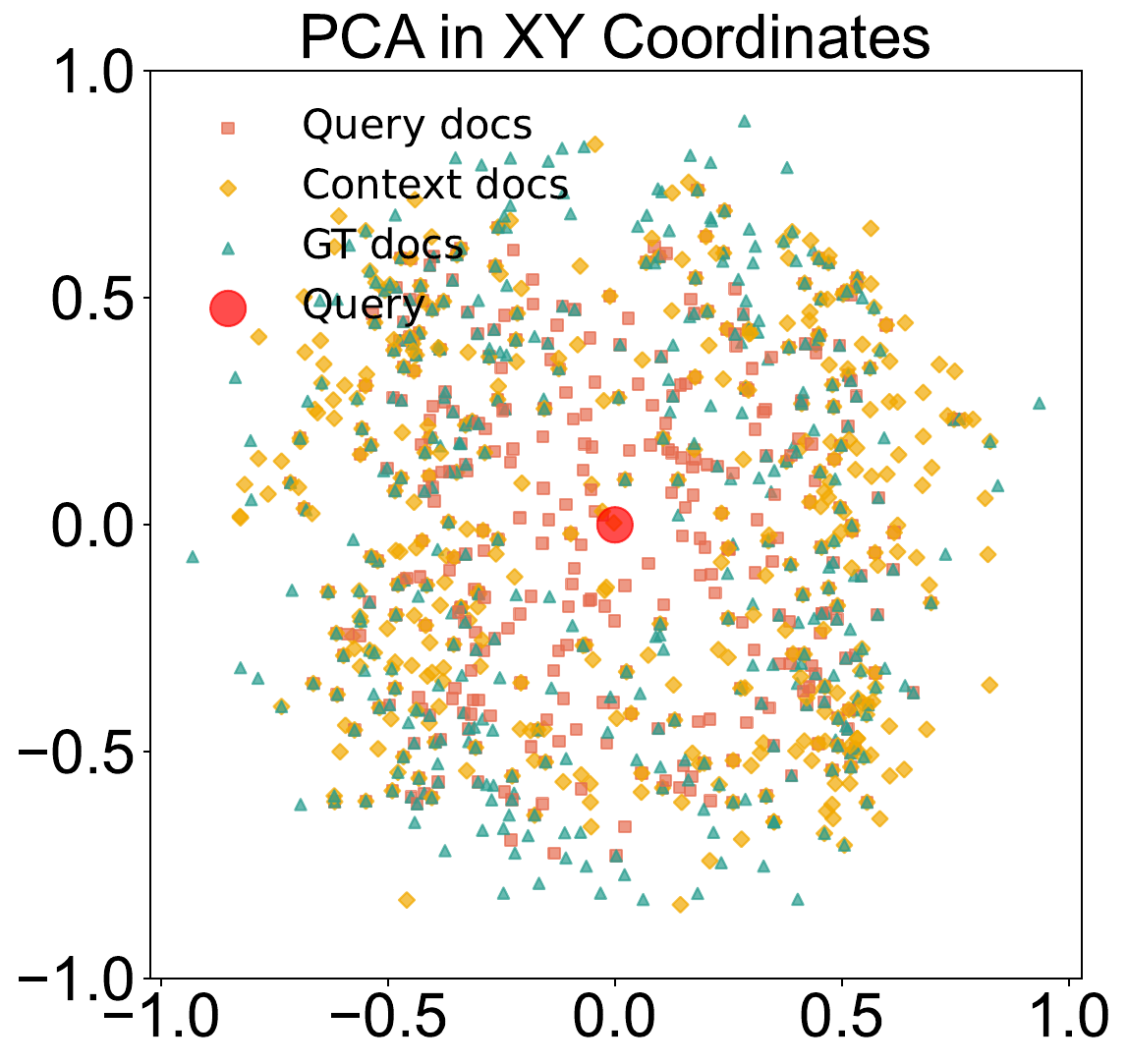}
    \caption{Relationship between query, context, and GT documents in the XY coordinate.}
    \label{fig: pca_xy}
\end{figure}

\begin{figure}[!ht]
    \centering
    \includegraphics[width=0.7\linewidth]{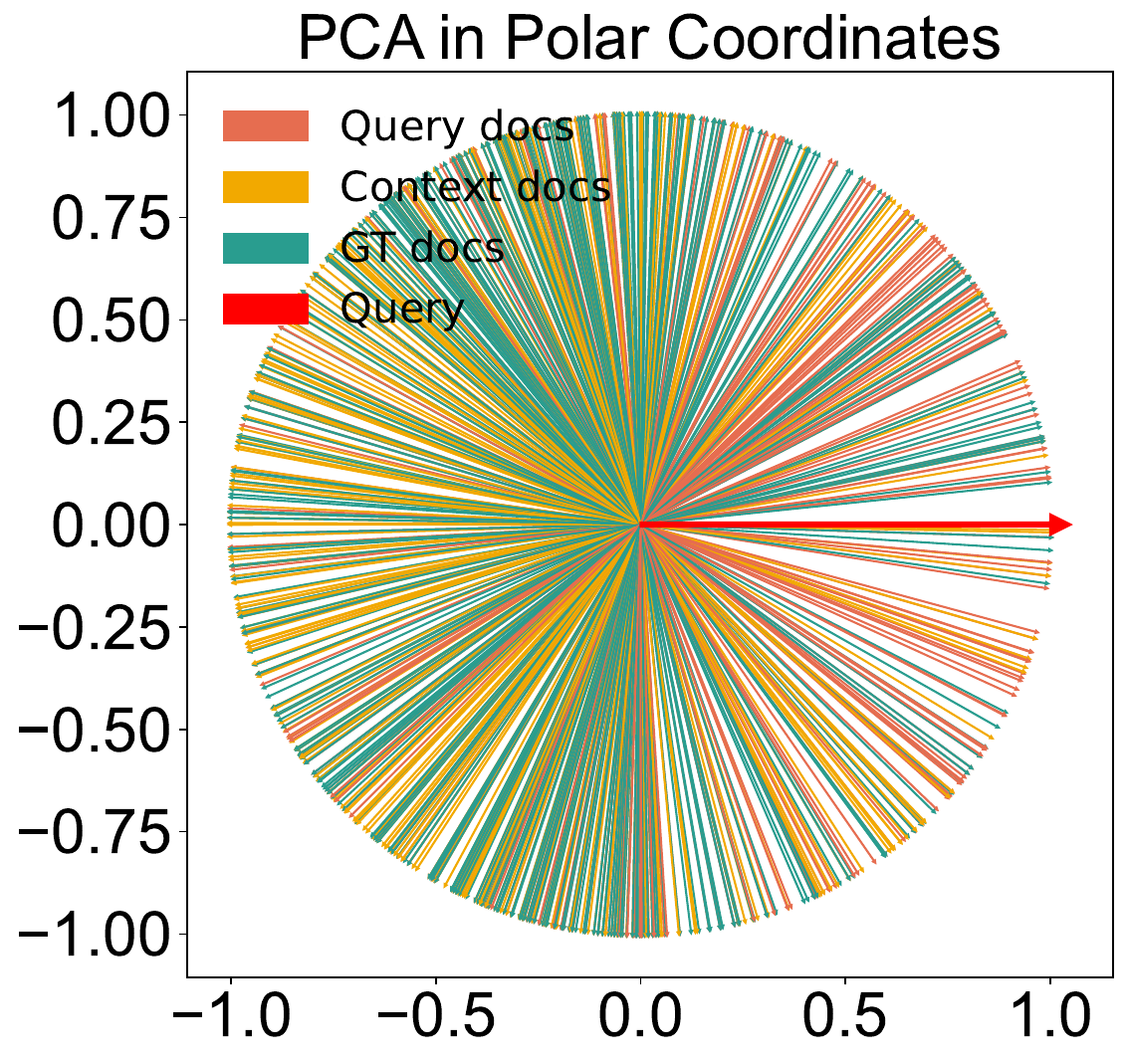}
    \caption{Relationship between query, context, and GT documents in the polar coordinate.}
    \label{fig: pca_polar}
\end{figure}

To figure out the relationship between the query, context, and GT documents, we projected these documents into the 2D space. Figs. \ref{fig: pca_xy} and \ref{fig: pca_polar} illustrate their relationship in xy and polar coordinates, respectively. Remind that we only visualize 200 samples, and we set the origin of the coordinates as the query. Most of the query documents are located besides the origin (i.e., the query), while the GT documents are kind of far from the origin, demonstrating that the GT documents may differ from the query documents in the vector space. However, the context documents can compensate for this gap as their distribution is similar to that of the GT documents when they are far away from the origin. These results further demonstrate that the LLM--generated pseudo-context can compensate for the sparse queries, enhancing the performance of RAG systems.

\section{Prompt template}
\label{seca: prompt}

In this section, we list all prompt templates used in this study.

\begin{figure*}[!ht]
\centering
\begin{tcolorbox}[colback=gray!5,colframe=black,title={Prompt for answer generation without retrievals}, width=\textwidth]

\{Question\}

Answer the question using a single word or phrase.

\end{tcolorbox}
\caption{Prompt used for generating final answers without retrievals.}
\end{figure*}

\begin{figure*}[!ht]
\centering
\begin{tcolorbox}[colback=gray!5,colframe=black,title={Prompt for answer generation with retrievals}, width=\textwidth]

\{Question\}

<Passage\_1>

<Passage\_2>

$\cdots$

<Passage\_$k$>

Answer the question based on the above context using a single word or phrase.

\end{tcolorbox}
\caption{Prompt used for generating final answers with retrievals.}
\end{figure*}

\begin{figure*}[!ht]
\centering
\begin{tcolorbox}[colback=gray!5,colframe=black,title={Prompt for pseudo context generation}, width=\textwidth]

\{Question\}

Write a passage to answer this question.

\end{tcolorbox}
\caption{Prompt used for generating pseudo context.}
\end{figure*}

\begin{figure*}[!ht]
\centering
\begin{tcolorbox}[colback=gray!5,colframe=black,title={Prompt for CoT generation}, width=\textwidth]

Answer the following question:

\{Question\}

Give the rationale before answering

\end{tcolorbox}
\caption{Prompt used for generating CoT.}
\end{figure*}

\begin{figure*}[!ht]
\centering
\begin{tcolorbox}[colback=gray!5,colframe=black,title={Prompt for retrieval judge}, width=\textwidth]

\{Question\}

Determine whether external information is needed to answer the question accurately. 
Respond with "Yes" if additional information is required, or "No" if the question can be answered without it.

\end{tcolorbox}
\caption{Prompt used for determining whether to retrieve.}
\end{figure*}

\section{Additional Uncertainty Measures}
\label{seca: additional_uncertainty_measure}

\begin{figure*}[!ht]
    \centering
    \includegraphics[width=1.0\linewidth]{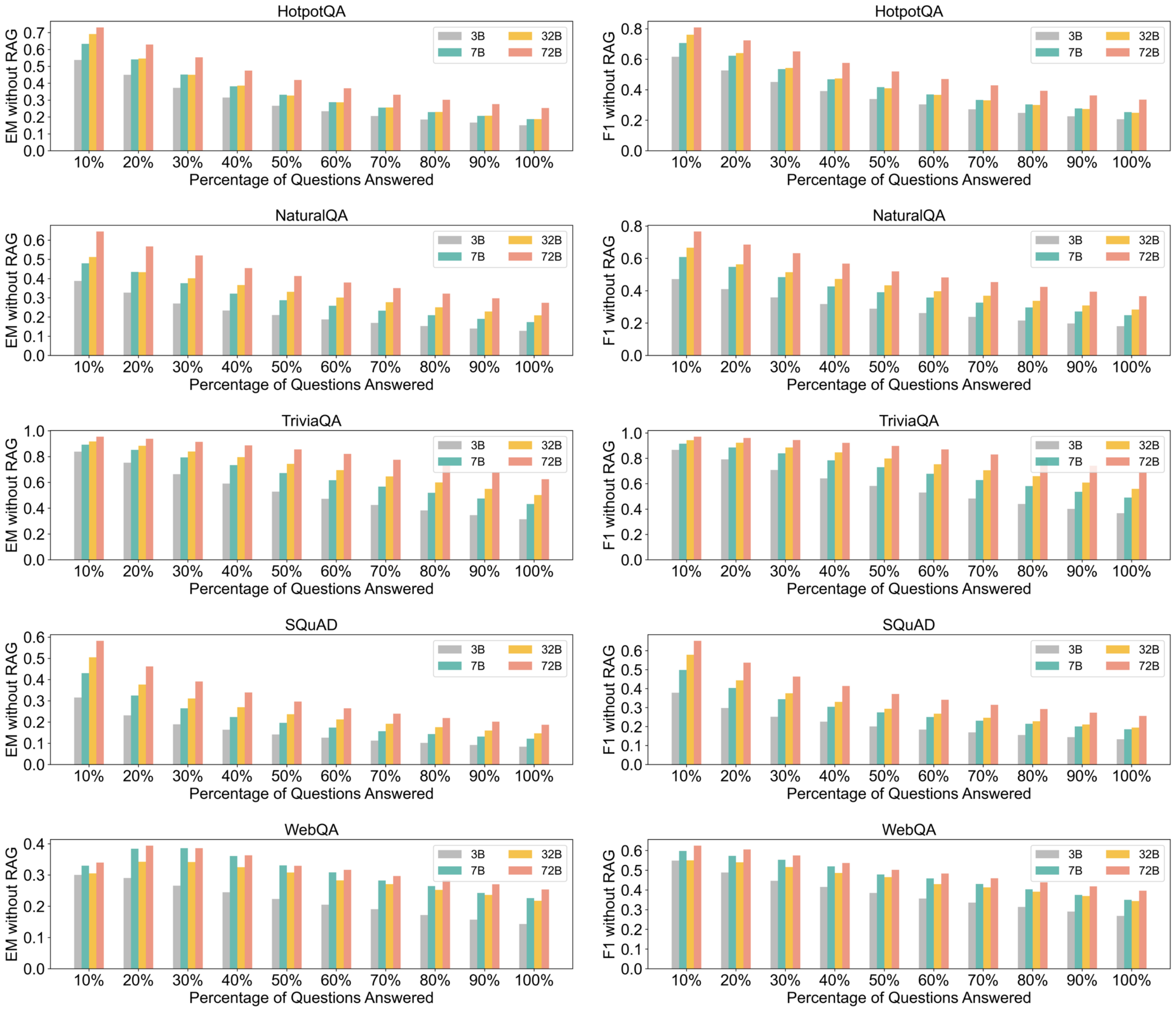}
    \caption{Evaluation results (EM and F1) across five datasets with \textit{Qwen2.5} series models as the generators. The percentage of questions is divided by the uncertainty scores of the generated answers. Without external retrievals, a lower uncertainty score can lead to higher generation accuracy.}
    \label{figa: u_percentage_EM_F1}
\end{figure*}

Beyond threshold-based analysis, we further investigate uncertainty by ranking queries according to their uncertainty scores and progressively selecting subsets of queries with the lowest uncertainty. Specifically, the \emph{query ratio} denotes the proportion of queries whose uncertainty is no greater than a given uncertainty threshold. Figure~\ref{figa: u_percentage_EM_F1} reports the Exact Match (EM) and F1 scores \emph{without retrieval} across five benchmark datasets as the query ratio increases from 10\% to 100\%. Across all datasets and model scales, a consistent trend emerges: queries with lower uncertainty achieve substantially higher EM and F1 scores, while including more high-uncertainty queries gradually degrades performance. This observation holds for both factoid (e.g., TriviaQA, SQuAD) and multi-hop or open-domain settings (e.g., HotpotQA, WebQA).

Moreover, larger models consistently dominate smaller ones at the same query ratio, indicating that stronger LLMs not only achieve higher parametric accuracy but also produce more reliable uncertainty estimates. Importantly, even when answering only a small fraction of low-uncertainty queries (e.g., top 20\%--30\%), the models can retain a large portion of their maximum achievable accuracy, suggesting that uncertainty effectively captures answer correctness.

\section{Additional evaluation results}
\label{seca: additional_results}

In this section, we report more evaluation results, including those with top-5 retrievals or with e5 as the retriever.

\subsection{Results with top-5 retrievals.}

\input{tables/results_RAG5}

Table~\ref{tab: results_RAG5} reports the results when increasing the number of retrieved passages to top-5, using \textit{Qwen2.5-7B-Instruct} as the generator. Overall, similar trends to the top-3 setting are observed, while performance is consistently improved across most methods due to the increased evidence coverage.

\paragraph{Overall performance.}
DTR remains the best-performing method under top-5 retrievals, achieving the highest average EM/F1 across all uncertainty thresholds. Compared with standard RAG, DTR improves the average EM/F1 from 37.30/47.60 to up to 38.78/49.28, demonstrating that uncertainty-guided triggering continues to be effective even when more retrieved passages are provided.

\paragraph{Effect of increased retrieval depth.}
While standard RAG benefits from additional retrieved passages, it also becomes more susceptible to noise introduced by less relevant evidence. On the contrary, DTR consistently yields higher accuracy, indicating that selectively triggering retrieval and adaptively leveraging retrieved information is crucial for mitigating noise in long-context settings.

\paragraph{Impact of uncertainty thresholds.}
Varying the uncertainty threshold $u$ controls the trade-off between retrieval frequency and accuracy. Smaller thresholds result in higher trigger ratios and slightly better performance on reasoning-intensive datasets such as HotpotQA and NaturalQA, whereas larger thresholds reduce retrieval usage with minimal performance degradation. These results further confirm the robustness of DTR across different retrieval depths.

Overall, the top-5 retrieval results reinforce our main findings: dynamic, uncertainty-guided triggering with dual-path retrieval remains effective and robust as the retrieval depth increases.

\subsection{Results with e5 as the retriever.}

\input{tables/results_e5}

Table~\ref{tab: results_e5} reports the main results when replacing \textit{bge} with \textit{e5} as the retriever, while keeping \textit{Qwen2.5-7B-Instruct} as the generator and using top-3 retrieved passages. Overall, the performance trends are consistent with those observed using \textit{bge}, demonstrating that DTR generalizes well across different retrieval models.

\paragraph{Overall performance.}
DTR consistently achieves the best or second-best performance across all datasets. In particular, DTR improves the average EM/F1 from 36.86/47.18 (standard RAG) to up to 38.42/48.86, confirming that uncertainty-guided dynamic triggering remains effective even with a different retriever.

\paragraph{Robustness across datasets.}
Across reasoning-intensive datasets such as HotpotQA and NaturalQA, DTR yields clear gains over standard RAG and retrieval-enhanced prompting methods. On TriviaQA and WebQA, where retrieval noise is more prominent, DTR maintains strong performance by selectively activating retrieval, highlighting its robustness to retriever variability.

\paragraph{Effect of uncertainty thresholds.}
As with \textit{bge}, varying the uncertainty threshold $u$ allows DTR to balance retrieval frequency and accuracy. Smaller thresholds lead to higher trigger ratios and slightly better overall performance, while larger thresholds reduce retrieval usage with minimal degradation. This further confirms that uncertainty-based triggering provides a stable and retriever-agnostic control mechanism.

\begin{figure*}[!ht]
    \centering
    \includegraphics[width=0.8\linewidth]{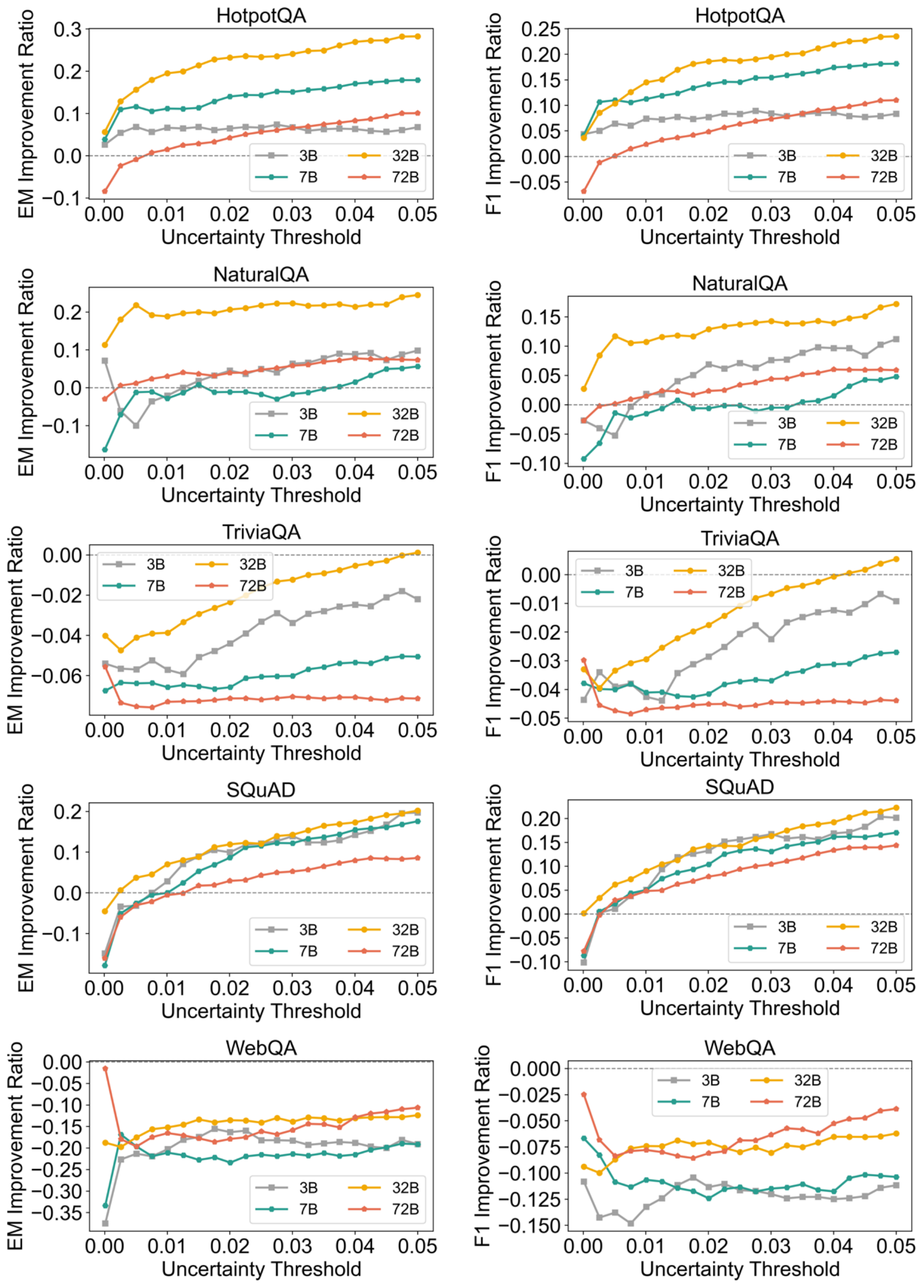}
    \caption{Uncertainty scaling results across various model sizes.}
    \label{figa: EM_F1_improvement_ratio}
\end{figure*}

\begin{figure*}[!ht]
    \centering
    \includegraphics[width=0.8\linewidth]{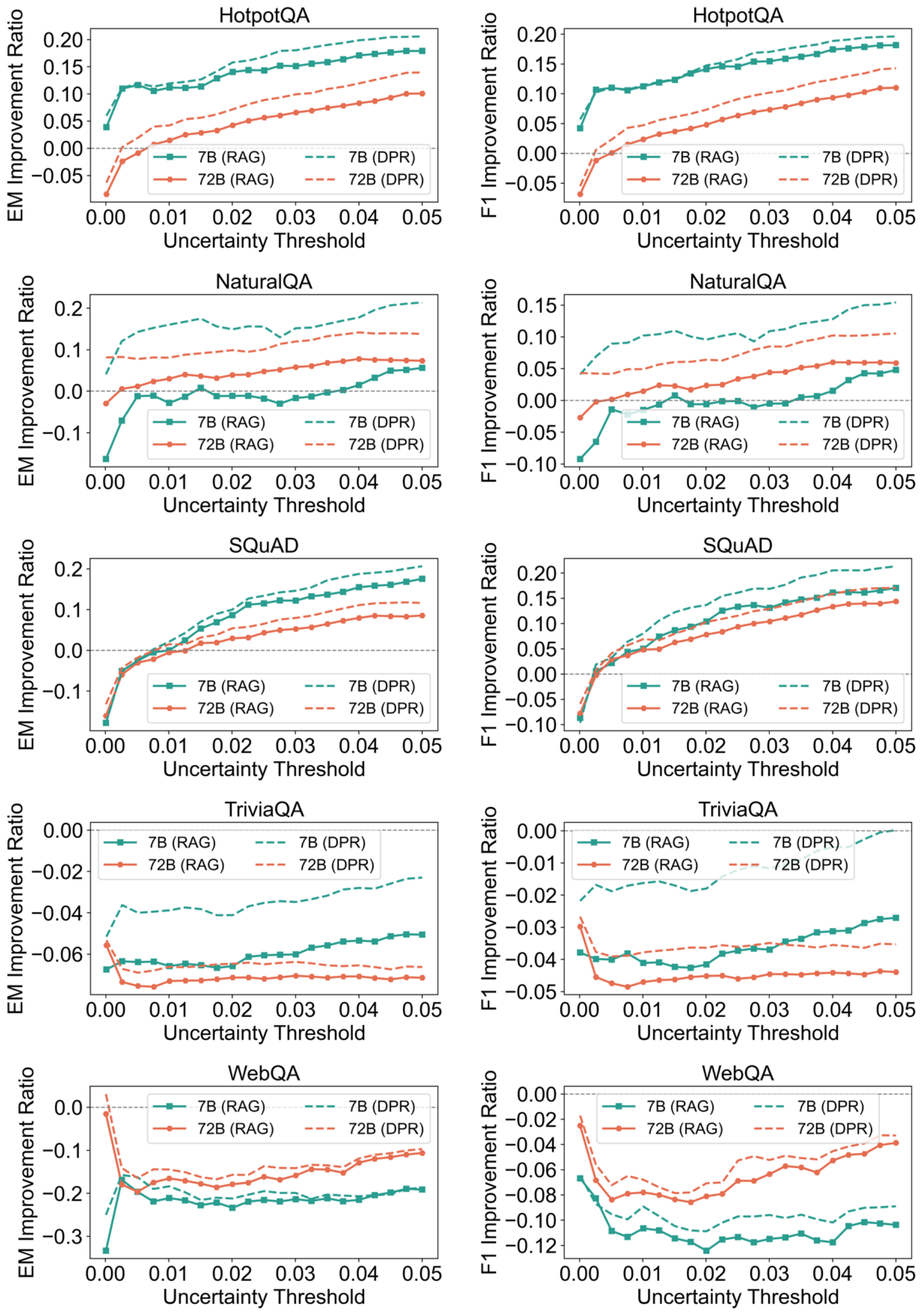}
    \caption{Uncertainty scaling results across different retrieval mechanisms.}
    \label{figa: EM_imp_compare_all}
\end{figure*}

Overall, the results with \textit{e5} demonstrate that DTR is not tied to a specific retriever and can consistently improve RAG performance across different retrieval backbones.

\section{Additional uncertainty scaling results}
\label{seca: uncertainty_scaling}

In this section, we present all uncertainty scaling results across five benchmarks, various model sizes, and different retrieval mechanisms, as shown in Figures~\ref{figa: EM_F1_improvement_ratio} and \ref{figa: EM_imp_compare_all}.

\section{Additional case studies}
\label{seca: additional_case_study}

\begin{figure*}[!ht]
    \centering
    \includegraphics[width=0.8\linewidth]{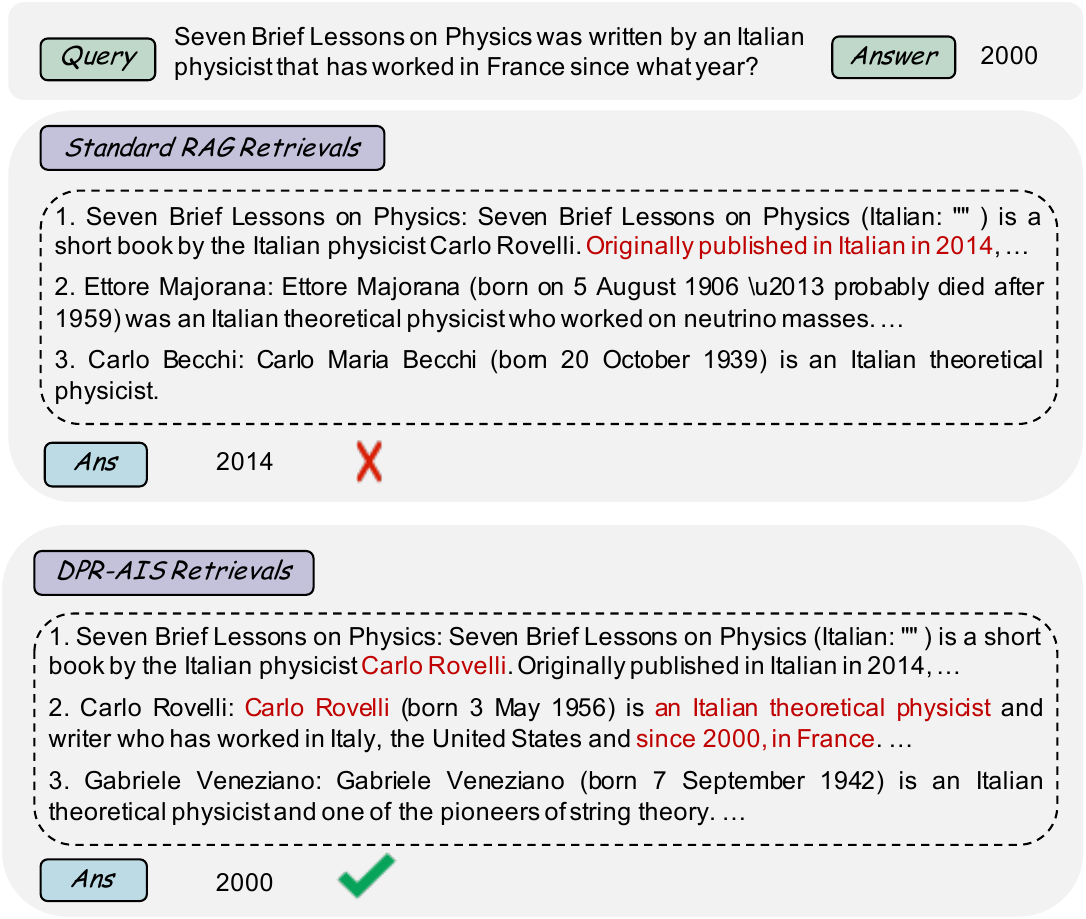}
    \caption{Comparison between standard RAG and DPR-AIS.}
    \label{figa: case_study_2}
\end{figure*}

Figure \ref{figa: case_study_2} compares the retrievals and answers achieved by standard RAG and DPR-AIS, respectively. Using the query to retrieve documents only can lead to similar but irrelevant results, which in turn results in a wrong final answer. However, our proposed dual-path retrieval mechanism can compensate for sparse queries and retrieve more relevant information, ultimately leading to accurate answers.

\section{Use of LLMs}

In the preparation of this paper, large language models (LLMs) were used solely for the purpose of polishing the writing, including grammar correction, improving sentence fluency, and ensuring a consistent academic tone. All core intellectual content—including the conceptualization of the proposed method, the design and execution of experiments, the analysis and interpretation of results, and the conclusions drawn—is the original work of the authors. The authors take full responsibility for the entire content of this paper, including any text generated with the assistance of LLMs.

\end{document}

%% file: tables/main_results.tex
\begin{table*}[!ht]
\centering
\resizebox{2.\columnwidth}{!}{
\begin{tabular}{l c c c c c c c c c c c c}

\toprule
\multirow{2}{*}{\textbf{Method}} & \multicolumn{2}{c}{\textbf{HotpotQA}} & \multicolumn{2}{c}{\textbf{NaturalQA}} & \multicolumn{2}{c}{\textbf{TriviaQA}} & \multicolumn{2}{c}{\textbf{SQuAD}} & \multicolumn{2}{c}{\textbf{WebQA}} & \multicolumn{2}{c}{\textbf{Average}}\\
 & \textbf{EM} & \textbf{F1} & \textbf{EM} & \textbf{F1} & \textbf{EM} & \textbf{F1} & \textbf{EM} & \textbf{F1} & \textbf{EM} & \textbf{F1} & \textbf{EM} & \textbf{F1}\\
\midrule
\multicolumn{13}{c}{\textit{Qwen2.5-7B-Instruct}} \\
\midrule
No Retrieval & 18.73 & 25.38 & 17.40 & 24.95 & 43.24 & 49.12 & 12.20 & 18.63 & 22.59 & 35.03 & 22.83 & 30.62\\
Standard RAG & \underline{36.18} & \underline{46.56} & 34.29 & 44.00 & 56.30 & 64.09 & 27.57 & 36.12 & 24.70 & 38.27 & 35.81 & 45.81\\
LLM Judge (7B) & 18.80 & 25.49 & 17.40 & 24.95 & 43.27 & 49.15 & 12.21 & 18.64 & 22.59 & 35.03 & 22.85 & 30.65\\
Trigger Ratio & \multicolumn{2}{c}{(0.5\%)} & \multicolumn{2}{c}{(0.0\%)} & \multicolumn{2}{c}{(0.1\%)} & \multicolumn{2}{c}{(0.0\%)} & \multicolumn{2}{c}{(0.0\%)} & \multicolumn{2}{c}{(0.1\%)}\\
LLM Judge (72B) & 34.56 & 44.60 & 26.79 & 35.64 & 49.92 & 56.38 & 25.14 & 33.21 & 23.67 & 36.23 & 32.02 & 41.21\\
Trigger Ratio & \multicolumn{2}{c}{(88.7\%)} & \multicolumn{2}{c}{(43.7\%)} & \multicolumn{2}{c}{(27.8\%)} & \multicolumn{2}{c}{(70.0\%)} & \multicolumn{2}{c}{(26.9\%)} & \multicolumn{2}{c}{(51.4\%)}\\
HyDE & 32.64 & 42.40 & 35.32 & 44.92 & 55.31 & 63.17 & 24.52 & 32.62 & 25.69 & 39.40 & 34.70 & 44.50\\
Q2D & 34.54 & 44.44 & 35.93 & 45.87 & 57.48 & 65.29 & 27.34 & 35.93 & 25.39 & 39.62 & 36.14 & 46.23\\
CoT & 34.44 & 44.57 & 36.18 & 45.89 & 57.39 & 65.26 & 27.09 & 35.57 & 25.84 & 39.99 & 36.19 & 46.25\\
\rowcolor{lightgray!30}
DTR ($u$ = 0.001) & \textbf{36.53} & \textbf{46.95} & \textbf{38.03} & \textbf{47.83} & 59.13 & 67.12 & \textbf{29.38} & \textbf{37.84} & 26.28 & 40.65 & \textbf{37.87} & \textbf{48.08}\\
Trigger Ratio & \multicolumn{2}{c}{(87.7\%)} & \multicolumn{2}{c}{(94.8\%)} & \multicolumn{2}{c}{(90.8\%)} & \multicolumn{2}{c}{(96.8\%)} & \multicolumn{2}{c}{(91.8\%)} & \multicolumn{2}{c}{(92.4\%)}\\
\rowcolor{lightgray!30}
DTR ($u$ = 0.005) & 36.16 & 46.55 & \underline{37.42} & \underline{47.35} & \underline{59.41} & \underline{67.23} & \underline{29.27} & \underline{37.69} & \underline{26.77} & \underline{41.17} & \underline{37.81} & \underline{48.00}\\
Trigger Ratio & \multicolumn{2}{c}{(83.5\%)} & \multicolumn{2}{c}{(90.4\%)} & \multicolumn{2}{c}{(83.0\%)} & \multicolumn{2}{c}{(93.6\%)} & \multicolumn{2}{c}{(84.5\%)} & \multicolumn{2}{c}{(87.0\%)}\\
\rowcolor{lightgray!30}
DTR ($u$ = 0.01) & 36.02 & 46.34 & 37.15 & 47.13 & \textbf{59.51} & \textbf{67.24} & 29.10 & 37.45 & \textbf{27.17} & \textbf{41.28} & 37.79 & 47.89\\
Trigger Ratio & \multicolumn{2}{c}{(81.0\%)} & \multicolumn{2}{c}{(87.5\%)} & \multicolumn{2}{c}{(78.7\%)} & \multicolumn{2}{c}{(91.2\%)} & \multicolumn{2}{c}{(81.1\%)} & \multicolumn{2}{c}{(83.9\%)}\\

\midrule
\multicolumn{13}{c}{\textit{Qwen2.5-72B-Instruct}} \\
\midrule

No Retrieval & 25.32 & 33.54 & 27.42 & 36.70 & 62.51 & 68.99 & 18.72 & 25.64 & 25.39 & 39.61 & 31.87 & 40.90\\
Standard RAG & 39.53 & 51.00 & 37.56 & 49.27 & 64.27 & 72.56 & 29.86 & 39.85 & 22.93 & 40.95 & 38.83 & 50.73\\
LLM Judge (7B) & 25.42 & 33.66 & 27.42 & 36.70 & 62.51 & 68.99 & 18.73 & 25.66 & 25.39 & 39.61 & 31.90 & 40.92\\
Trigger Ratio & \multicolumn{2}{c}{(0.5\%)} & \multicolumn{2}{c}{(0.0\%)} & \multicolumn{2}{c}{(0.1\%)} & \multicolumn{2}{c}{(0.0\%)} & \multicolumn{2}{c}{(0.0\%)} & \multicolumn{2}{c}{(0.1\%)}\\
LLM Judge (72B) & 39.24 & 50.54 & 33.91 & 44.90 & 65.51 & 72.61 & 28.52 & 37.63 & \underline{25.54} & 40.78 & 38.54 & 49.29\\
Trigger Ratio & \multicolumn{2}{c}{(88.7\%)} & \multicolumn{2}{c}{(43.7\%)} & \multicolumn{2}{c}{(27.8\%)} & \multicolumn{2}{c}{(70.0\%)} & \multicolumn{2}{c}{(26.9\%)} & \multicolumn{2}{c}{(51.4\%)}\\
HyDE & 35.08 & 45.62 & 37.40 & 49.13 & 61.42 & 69.71 & 26.03 & 35.66 & 24.51 & 42.12 & 36.89 & 48.45\\
Q2D & 36.95 & 47.93 & 37.73 & 49.72 & 63.28 & 71.71 & 29.09 & 38.98 & 23.97 & 41.70 & 38.20 & 50.01\\
CoT & 36.87 & 47.96 & 37.56 & 49.54 & 64.02 & 72.47 & 29.22 & 38.92 & 24.06 & 42.12 & 38.35 & 50.20\\
\rowcolor{lightgray!30}
DTR ($u$ = 0.001) & \textbf{40.68} & \textbf{52.28} & \textbf{38.92} & \textbf{51.28} & 66.15 & 73.98 & \textbf{31.04} & \textbf{40.97} & 24.56 & 42.26 & 40.27 & \textbf{52.16}\\
Trigger Ratio & \multicolumn{2}{c}{(83.9\%)} & \multicolumn{2}{c}{(88.1\%)} & \multicolumn{2}{c}{(71.0\%)} & \multicolumn{2}{c}{(91.9\%)} & \multicolumn{2}{c}{(82.7\%)} & \multicolumn{2}{c}{(83.5\%)}\\
\rowcolor{lightgray!30}
DTR ($u$ = 0.005) & \underline{40.27} & \underline{51.74} & \underline{38.56} & \underline{50.93} & \underline{67.06} & \underline{74.61} & \underline{30.75} & \underline{40.49} & \textbf{25.64} & \underline{42.95} & \textbf{40.46} & \underline{52.14}\\
Trigger Ratio & \multicolumn{2}{c}{(77.1\%)} & \multicolumn{2}{c}{(79.2\%)} & \multicolumn{2}{c}{(59.8\%)} & \multicolumn{2}{c}{(85.6\%)} & \multicolumn{2}{c}{(73.4\%)} & \multicolumn{2}{c}{(75.0\%)}\\
\rowcolor{lightgray!30}
DTR ($u$ = 0.01) & 39.89 & 51.27 & 38.42 & 50.71 & \textbf{67.24} & \textbf{74.72} & 30.51 & 40.15 & \textbf{25.64} & \textbf{43.01} & \underline{40.34} & 51.97\\
Trigger Ratio & \multicolumn{2}{c}{(72.7\%)} & \multicolumn{2}{c}{(74.5\%)} & \multicolumn{2}{c}{(54.4\%)} & \multicolumn{2}{c}{(81.5\%)} & \multicolumn{2}{c}{(69.5\%)} & \multicolumn{2}{c}{(70.5\%)}\\

\bottomrule

\end{tabular}}
\caption{Main results across five QA datasets with \textit{Qwen2.5-7B-Instruct} or \textit{Qwen2.5-72B-Instruct} as the generator and \textit{bge} as the retriever, respectively. Except for experiments that require no retrievals, all results are generated based on \textit{top-3 retrievals}. \textbf{Trigger Ratio} indicates the proportion of queries for which the retriever was triggered. \textbf{Bold} and \underline{underlined} values represent the best and second-best scores, respectively.}
\label{tab: results}
\end{table*}

%% file: tables/ablation_results.tex
\begin{table*}[!ht]
\centering
\resizebox{2.\columnwidth}{!}{
\begin{tabular}{l c c c c c c c c c c c c}

\toprule
\multirow{2}{*}{\textbf{Method}} & \multicolumn{2}{c}{\textbf{HotpotQA}} & \multicolumn{2}{c}{\textbf{NaturalQA}} & \multicolumn{2}{c}{\textbf{TriviaQA}} & \multicolumn{2}{c}{\textbf{SQuAD}} & \multicolumn{2}{c}{\textbf{WebQA}} & \multicolumn{2}{c}{\textbf{Average}}\\
 & \textbf{EM} & \textbf{F1} & \textbf{EM} & \textbf{F1} & \textbf{EM} & \textbf{F1} & \textbf{EM} & \textbf{F1} & \textbf{EM} & \textbf{F1} & \textbf{EM} & \textbf{F1}\\
\midrule
\multicolumn{13}{c}{\textit{Qwen2.5-7B-Instruct}} \\
\midrule
\rowcolor{lightgray!30}
DTR ($u$ = 0.001) & \underline{36.53} & \underline{46.95} & \underline{38.03} & \underline{47.83} & \textbf{59.13} & \textbf{67.12} & \textbf{29.38} & \textbf{37.84} & \textbf{26.28} & \textbf{40.65} & \textbf{37.87} & \underline{48.08}\\
\textit{w/o} UGT & \textbf{37.27} & \textbf{47.71} & \textbf{38.09} & \textbf{47.88} & \underline{58.82} & \underline{66.94} & \underline{29.19} & \underline{37.82} & \underline{25.84} & \underline{40.30} & \underline{37.84} & \textbf{48.13}\\
\textit{w/o} DPR & 35.46 & 45.77 & 34.76 & 44.41 & 56.80 & 64.44 & 27.80 & 36.20 & 25.10 & 38.49 & 35.98 & 45.86\\
\textit{w/o} AIS (1$q$ + 2$p$) & 30.38 & 40.20 & 30.36 & 39.18 & 51.90 & 59.16 & 22.15 & 29.54 & 20.96 & 33.53 & 31.15 & 40.32\\
\textit{w/o} AIS (2$q$ + 1$p$) & 34.30 & 44.24 & 33.30 & 42.75 & 55.10 & 62.56 & 26.08 & 34.05 & 23.28 & 36.93 & 34.41 & 44.10\\

\midrule
\multicolumn{13}{c}{\textit{Qwen2.5-72B-Instruct}} \\
\midrule

\rowcolor{lightgray!30}
DTR ($u$ = 0.001) & \textbf{40.68} & \textbf{52.28} & \underline{38.92} & \underline{51.28} & \textbf{66.15} & \textbf{73.98} & \textbf{31.04} & \textbf{40.97} & \textbf{24.56} & \textbf{42.26} & \textbf{40.27} & \textbf{52.16}\\
\textit{w/o} UGT & \underline{40.54} & \underline{52.12} & \textbf{39.47} & \textbf{51.51} & 64.62 & 73.16 & \underline{30.63} & \underline{40.85} & 23.92 & \underline{41.82} & \underline{39.84} & \underline{51.89}\\
\textit{w/o} DPR & 39.99 & 51.44 & 37.56 & 49.40 & \underline{66.03} & \underline{73.66} & 30.37 & 40.06 & \underline{23.97} & 41.58 & 39.58 & 51.23\\

\bottomrule

\end{tabular}}
\caption{Ablation results across five QA datasets with \textit{Qwen2.5-7B-Instruct} or \textit{Qwen2.5-72B-Instruct} as the generator. \textit{w/o} AIS (w/ 1$q$ + 2$p$) and \textit{w/o} AIS (w/ 2$q$ + 1$p$) denote retrieving documents using fixed proportions, i.e., two from the query and one from the pseudo-context, or vice versa. \textbf{Bold} and \underline{underlined} values represent the best and second-best scores, respectively.}
\label{tab: ablation_results}
\end{table*}

%% file: tables/retrieval_acc.tex




\begin{table}[!ht] 
\centering 
\begin{tabular}{l c c}
\toprule 

\multirow{3}{*}{Method} & \multicolumn{2}{c}{HotpotQA} \\

& \textit{bge} & \textit{e5} \\

\midrule
Standard RAG & \underline{61.9\%} & \underline{59.3\%} \\
HyDE & 49.9\% & 49.9\% \\
Q2D & 54.1\% & 55.2\%  \\
CoT & 53.6\% & 54.7\%  \\
DPR & \textbf{62.7\%} & \textbf{62.6\%} \\
\bottomrule
\end{tabular}
\caption{Comparison of retrieval performance between our proposed dual-path retrieval method and the other query expansion methods. The measurement metric is \textit{recall@3}.}
\label{tab: retrieval_acc}
\end{table}

%% file: tables/results_RAG5.tex
\begin{table*}[!ht]
\centering
\resizebox{2.\columnwidth}{!}{
\begin{tabular}{l c c c c c c c c c c c c}

\toprule
\multirow{2}{*}{\textbf{Method}} & \multicolumn{2}{c}{\textbf{HotpotQA}} & \multicolumn{2}{c}{\textbf{NaturalQA}} & \multicolumn{2}{c}{\textbf{TriviaQA}} & \multicolumn{2}{c}{\textbf{SQuAD}} & \multicolumn{2}{c}{\textbf{WebQA}} & \multicolumn{2}{c}{\textbf{Average}}\\
 & \textbf{EM} & \textbf{F1} & \textbf{EM} & \textbf{F1} & \textbf{EM} & \textbf{F1} & \textbf{EM} & \textbf{F1} & \textbf{EM} & \textbf{F1} & \textbf{EM} & \textbf{F1}\\
\midrule
No Retrieval & 18.73 & 25.38 & 17.40 & 24.95 & 43.24 & 49.12 & 12.20 & 18.63 & 22.59 & 35.03 & 22.83 & 30.62\\
Standard RAG & \underline{37.54} & \textbf{48.40} & 36.45 & 46.23 & 57.33 & 65.21 & 30.01 & 38.84 & 25.15 & 39.29 & 37.30 & 47.60\\
LLM Judge (7B) & 18.77 & 25.48 & 17.40 & 24.95 & 43.27 & 49.15 & 12.21 & 18.64 & 22.59 & 35.03 & 22.85 & 30.65\\
Trigger Ratio & \multicolumn{2}{c}{(0.5\%)} & \multicolumn{2}{c}{(0.0\%)} & \multicolumn{2}{c}{(0.1\%)} & \multicolumn{2}{c}{(0.0\%)} & \multicolumn{2}{c}{(0.0\%)} & \multicolumn{2}{c}{(0.1\%)}\\
LLM Judge (72B) & 35.80 & 46.32 & 27.89 & 36.78 & 50.45 & 56.92 & 26.98 & 35.21 & 24.11 & 36.73 & 33.05 & 42.39\\
Trigger Ratio & \multicolumn{2}{c}{(88.7\%)} & \multicolumn{2}{c}{(43.7\%)} & \multicolumn{2}{c}{(27.8\%)} & \multicolumn{2}{c}{(70.0\%)} & \multicolumn{2}{c}{(26.9\%)} & \multicolumn{2}{c}{(51.4\%)}\\
HyDE & 34.17 & 43.86 & 36.70 & 46.35 & 56.25 & 64.15 & 26.27 & 34.83 & 27.07 & 41.00 & 36.09 & 46.04\\
Q2D & 35.81 & 45.89 & 36.87 & 47.15 & 58.48 & 66.34 & 29.40 & 38.39 & 26.03 & 40.18 & 37.32 & 47.59\\
CoT & 35.92 & 45.94 & 37.17 & 47.46 & 58.61 & 66.42 & 29.21 & 37.89 & 26.18 & 40.54 & 37.42 & 47.65\\
\rowcolor{lightgray!30}
DTR ($u$ = 0.001) & \textbf{37.61} & \underline{48.30} & \textbf{38.50} & \textbf{48.63} & 60.30 & 68.29 & \textbf{30.99} & \textbf{39.92} & 26.48 & 41.23 & \textbf{38.78} & \textbf{49.28}\\
Trigger Ratio & \multicolumn{2}{c}{(87.7\%)} & \multicolumn{2}{c}{(94.8\%)} & \multicolumn{2}{c}{(90.8\%)} & \multicolumn{2}{c}{(96.8\%)} & \multicolumn{2}{c}{(91.8\%)} & \multicolumn{2}{c}{(92.4\%)}\\
\rowcolor{lightgray!30}
DTR ($u$ = 0.005) & 37.25 & 47.88 & \underline{37.95} & \underline{48.13} & \underline{60.55} & \underline{68.36} & \underline{30.85} & \underline{39.76} & \underline{27.21} & \underline{41.90} & \underline{38.76} & \underline{49.20}\\
Trigger Ratio & \multicolumn{2}{c}{(83.5\%)} & \multicolumn{2}{c}{(90.4\%)} & \multicolumn{2}{c}{(83.0\%)} & \multicolumn{2}{c}{(93.6\%)} & \multicolumn{2}{c}{(84.5\%)} & \multicolumn{2}{c}{(87.0\%)}\\
\rowcolor{lightgray!30}
DTR ($u$ = 0.01) & 37.12 & 47.70 & 37.67 & 47.95 & \textbf{60.70} & \textbf{68.39} & 30.69 & 39.53 & \textbf{27.56} & \textbf{42.10} & 38.75 & 49.13\\
Trigger Ratio & \multicolumn{2}{c}{(81.0\%)} & \multicolumn{2}{c}{(87.5\%)} & \multicolumn{2}{c}{(78.7\%)} & \multicolumn{2}{c}{(91.2\%)} & \multicolumn{2}{c}{(81.1\%)} & \multicolumn{2}{c}{(83.9\%)}\\
\bottomrule

\end{tabular}}
\caption{Main results across five QA datasets with \textit{Qwen2.5-7B-Instruct} as the generator and \textit{bge} as the retriever, respectively. Except for experiments that require no retrievals, all results are generated based on \textit{top-5 retrievals}. \textbf{Trigger Ratio} indicates the proportion of queries for which the retriever was triggered. \textbf{Bold} and \underline{underlined} values represent the best and second-best scores, respectively.}
\label{tab: results_RAG5}
\end{table*}

%% file: tables/results_e5.tex
\begin{table*}[!ht]
\centering
\resizebox{2.\columnwidth}{!}{
\begin{tabular}{l c c c c c c c c c c c c}

\toprule
\multirow{2}{*}{\textbf{Method}} & \multicolumn{2}{c}{\textbf{HotpotQA}} & \multicolumn{2}{c}{\textbf{NaturalQA}} & \multicolumn{2}{c}{\textbf{TriviaQA}} & \multicolumn{2}{c}{\textbf{SQuAD}} & \multicolumn{2}{c}{\textbf{WebQA}} & \multicolumn{2}{c}{\textbf{Average}}\\
 & \textbf{EM} & \textbf{F1} & \textbf{EM} & \textbf{F1} & \textbf{EM} & \textbf{F1} & \textbf{EM} & \textbf{F1} & \textbf{EM} & \textbf{F1} & \textbf{EM} & \textbf{F1}\\
\midrule
No Retrieval & 18.73 & 25.38 & 17.40 & 24.95 & 43.24 & 49.12 & 12.20 & 18.63 & 22.59 & 35.03 & 22.83 & 30.62\\
Standard RAG & 35.18 & 45.77 & 37.56 & 47.29 & 59.00 & 66.99 & 28.84 & 37.43 & 23.72 & 38.43 & 36.86 & 47.18\\
LLM Judge (7B) & 18.77 & 25.46 & 17.40 & 24.95 & 43.27 & 49.15 & 12.21 & 18.65 & 22.59 & 35.03 & 22.85 & 30.65\\
Trigger Ratio & \multicolumn{2}{c}{(0.5\%)} & \multicolumn{2}{c}{(0.0\%)} & \multicolumn{2}{c}{(0.1\%)} & \multicolumn{2}{c}{(0.0\%)} & \multicolumn{2}{c}{(0.0\%)} & \multicolumn{2}{c}{(0.1\%)}\\
LLM Judge (72B) & 33.60 & 43.83 & 28.53 & 37.66 & 51.26 & 57.82 & 26.37 & 34.44 & 24.16 & 36.93 & 32.78 & 42.14\\
Trigger Ratio & \multicolumn{2}{c}{(88.7\%)} & \multicolumn{2}{c}{(43.7\%)} & \multicolumn{2}{c}{(27.8\%)} & \multicolumn{2}{c}{(70.0\%)} & \multicolumn{2}{c}{(26.9\%)} & \multicolumn{2}{c}{(51.4\%)}\\
HyDE & 32.64 & 42.40 & 35.32 & 44.92 & 55.31 & 63.17 & 24.52 & 32.62 & 25.69 & 39.40 & 34.70 & 44.50\\
Q2D & 34.54 & 44.44 & 35.93 & 45.87 & 57.48 & 65.29 & 27.34 & 35.93 & 25.39 & 39.62 & 36.14 & 46.23\\
CoT & 34.44 & 44.57 & 36.18 & 45.89 & 57.39 & 65.26 & 27.09 & 35.57 & 25.84 & 39.99 & 36.19 & 46.25\\
\rowcolor{lightgray!30}
DTR ($u$ = 0.001) & \textbf{36.73} & \textbf{47.34} & \textbf{38.84} & \textbf{49.22} & 61.26 & 69.40 & \underline{29.11} & \textbf{37.68} & 26.18 & 40.65 & \textbf{38.42} & \textbf{48.86}\\
Trigger Ratio & \multicolumn{2}{c}{(87.7\%)} & \multicolumn{2}{c}{(94.8\%)} & \multicolumn{2}{c}{(90.8\%)} & \multicolumn{2}{c}{(96.8\%)} & \multicolumn{2}{c}{(91.8\%)} & \multicolumn{2}{c}{(92.4\%)}\\
\rowcolor{lightgray!30}
DTR ($u$ = 0.005) & \underline{36.29} & \underline{46.87} & \underline{38.12} & \underline{48.64} & \underline{61.54} & \textbf{69.49} & \textbf{29.13} & \underline{37.65} & \underline{26.97} & \underline{41.41} & 38.41 & \underline{48.81}\\
Trigger Ratio & \multicolumn{2}{c}{(83.5\%)} & \multicolumn{2}{c}{(90.4\%)} & \multicolumn{2}{c}{(83.0\%)} & \multicolumn{2}{c}{(93.6\%)} & \multicolumn{2}{c}{(84.5\%)} & \multicolumn{2}{c}{(87.0\%)}\\
\rowcolor{lightgray!30}
DTR ($u$ = 0.01) & 36.21 & 46.71 & 37.92 & 48.45 & \textbf{61.61} & \underline{69.46} & 29.02 & 37.47 & \textbf{27.36} & \textbf{41.57} & \underline{38.42} & 48.73\\
Trigger Ratio & \multicolumn{2}{c}{(81.0\%)} & \multicolumn{2}{c}{(87.5\%)} & \multicolumn{2}{c}{(78.7\%)} & \multicolumn{2}{c}{(91.2\%)} & \multicolumn{2}{c}{(81.1\%)} & \multicolumn{2}{c}{(83.9\%)}\\
\bottomrule

\end{tabular}}
\caption{Main results across five QA datasets with \textit{Qwen2.5-7B-Instruct} as the generator and \textit{e5} as the retriever, respectively. Except for experiments that require no retrievals, all results are generated based on \textit{top-3 retrievals}. \textbf{Trigger Ratio} indicates the proportion of queries for which the retriever was triggered. \textbf{Bold} and \underline{underlined} values represent the best and second-best scores, respectively.}
\label{tab: results_e5}
\end{table*}